\documentclass{article}

\usepackage{full/arxiv}

\usepackage{iftex}
\ifPDFTeX
  \usepackage[utf8]{inputenc} 
  \usepackage[T1]{fontenc}    
\else
  \usepackage{fontspec}       
\fi
\usepackage{hyperref}       
\hypersetup{hidelinks,pdftitle={What Does Multi-Agent Debate Actually Change?},
            pdfauthor={Chen Qian}}
\usepackage{xurl}            
\usepackage{booktabs}       
\usepackage{array}
\usepackage{amsmath,amssymb}
\usepackage{graphicx}
\usepackage{placeins}      
\usepackage{natbib}
\usepackage{fancyhdr}
\usepackage{tikz}
\usepackage{fvextra}       
\DefineVerbatimEnvironment{promptbox}{Verbatim}{fontsize=\scriptsize,breaklines,breakanywhere,frame=single,framesep=2mm}

\definecolor{figblue}{HTML}{2A78D6}
\definecolor{figorange}{HTML}{EB6834}
\definecolor{figaqua}{HTML}{1BAF7A}
\definecolor{figmuted}{HTML}{898781}
\definecolor{figink}{HTML}{0B0B0B}

\makeatletter\providecommand{\rd}[1]{\@ifundefined{rd@#1}{\textbf{??#1??}\PackageWarning{rd}{Undefined key #1}}{\csname rd@#1\endcsname}}\makeatother
\expandafter\def\csname rd@A/dyad/mixed/gap\endcsname{61.0}
\expandafter\def\csname rd@A/dyad/mixed/hostile/ld\endcsname{43.7}
\expandafter\def\csname rd@A/dyad/replies\endcsname{900}
\expandafter\def\csname rd@A/dyad/same/gap\endcsname{75.7}
\expandafter\def\csname rd@A/dyad/same/hostile/la\endcsname{71.3}
\expandafter\def\csname rd@A/dyad/same/hostile/ld\endcsname{12.0}
\expandafter\def\csname rd@A/friendly/sa/max\endcsname{96.3}
\expandafter\def\csname rd@A/friendly/sa/min\endcsname{70.7}
\expandafter\def\csname rd@A/hostile/sa/max\endcsname{19.3}
\expandafter\def\csname rd@A/hostile/sa/min\endcsname{9.5}
\expandafter\def\csname rd@A/neutral/sa/max\endcsname{92.4}
\expandafter\def\csname rd@A/neutral/sa/min\endcsname{64.7}
\expandafter\def\csname rd@A/replies/total\endcsname{7{,}797}
\expandafter\def\csname rd@A/sd/total\endcsname{9}
\expandafter\def\csname rd@A/triad/mixed/gap\endcsname{61.3}
\expandafter\def\csname rd@A/triad/mixed/hostile/ld\endcsname{38.5}
\expandafter\def\csname rd@A/triad/replies\endcsname{1{,}699}
\expandafter\def\csname rd@A/triad/same/gap\endcsname{82.4}
\expandafter\def\csname rd@A/triad/same/hostile/la\endcsname{77.6}
\expandafter\def\csname rd@A/triad/same/hostile/ld\endcsname{8.4}
\expandafter\def\csname rd@B/dyad/mixed/judgemoredisagree\endcsname{9.9}
\expandafter\def\csname rd@B/dyad/mixed/kappa\endcsname{0.72}
\expandafter\def\csname rd@B/dyad/mixed/ldtosd\endcsname{21.4}
\expandafter\def\csname rd@B/dyad/same/judgemoreagree\endcsname{7.2}
\expandafter\def\csname rd@B/dyad/same/judgemoredisagree\endcsname{1.5}
\expandafter\def\csname rd@B/dyad/same/n\endcsname{667}
\expandafter\def\csname rd@B/dyad/same/satosa\endcsname{99.8}
\expandafter\def\csname rd@B/exact/max\endcsname{92.2}
\expandafter\def\csname rd@B/exact/min\endcsname{81.8}
\expandafter\def\csname rd@B/kappa/max\endcsname{0.85}
\expandafter\def\csname rd@B/kappa/min\endcsname{0.71}
\expandafter\def\csname rd@B/latosa/max\endcsname{27.7}
\expandafter\def\csname rd@B/latosa/min\endcsname{19.1}
\expandafter\def\csname rd@B/triad/mixed/judgemoredisagree\endcsname{9.6}
\expandafter\def\csname rd@B/triad/mixed/kappa\endcsname{0.71}
\expandafter\def\csname rd@B/triad/mixed/n\endcsname{731}
\expandafter\def\csname rd@B/triad/mixed/satosa\endcsname{97.5}
\expandafter\def\csname rd@B/triad/same/judgemoreagree\endcsname{7.1}
\expandafter\def\csname rd@B/triad/same/judgemoredisagree\endcsname{0.7}
\expandafter\def\csname rd@B/triad/same/ldtosd\endcsname{7.0}
\expandafter\def\csname rd@B/triad/same/n\endcsname{731}
\expandafter\def\csname rd@C/family/ld/la/option\endcsname{53.8}
\expandafter\def\csname rd@C/family/ld/sa/option\endcsname{100.0}
\expandafter\def\csname rd@C/family/option/all\endcsname{5.2}
\expandafter\def\csname rd@C/ld/ld/option/max\endcsname{3.1}
\expandafter\def\csname rd@C/mixed/ld/la/option\endcsname{17.5}
\expandafter\def\csname rd@C/mixed/ld/sa/option\endcsname{55.0}
\expandafter\def\csname rd@C/mixed/option/all\endcsname{4.8}
\expandafter\def\csname rd@C/same/ld/sa/option\endcsname{100.0}
\expandafter\def\csname rd@C/same/option/all\endcsname{4.1}
\expandafter\def\csname rd@D/inkling/ld/ld/d1_pp\endcsname{$-$6.0}
\expandafter\def\csname rd@D/qwen/ld/ld/d1_pp\endcsname{$-$0.3}
\expandafter\def\csname rd@D/qwen/ld/ld/d2_argument\endcsname{5.81}
\expandafter\def\csname rd@D/qwen/ld/ld/d2_filler\endcsname{6.74}
\expandafter\def\csname rd@D/qwen/sa/sa/d2_argument\endcsname{6.86}
\expandafter\def\csname rd@D/qwen/sa/sa/d2_filler\endcsname{6.84}
\expandafter\def\csname rd@D1/inkling/first/after\endcsname{98.7}
\expandafter\def\csname rd@D1/inkling/first/before\endcsname{77.7}
\expandafter\def\csname rd@D1/inkling/later/after\endcsname{96.9}
\expandafter\def\csname rd@D1/inkling/later/before\endcsname{96.3}
\expandafter\def\csname rd@D1/inkling/later/change\endcsname{+0.6}
\expandafter\def\csname rd@D1/inkling/later_pushback/change\endcsname{$-$1.3}
\expandafter\def\csname rd@D1/qwen/first/after\endcsname{93.0}
\expandafter\def\csname rd@D1/qwen/first/before\endcsname{45.9}
\expandafter\def\csname rd@D1/qwen/later/after\endcsname{95.5}
\expandafter\def\csname rd@D1/qwen/later/before\endcsname{92.1}
\expandafter\def\csname rd@D1/qwen/later/change\endcsname{+3.4}
\expandafter\def\csname rd@D1/qwen/later_pushback/change\endcsname{+3.0}
\expandafter\def\csname rd@D1/qwen/later_pushback/down\endcsname{10}
\expandafter\def\csname rd@D1/qwen/later_pushback/n\endcsname{42}
\expandafter\def\csname rd@D1/qwen/later_pushback/up\endcsname{16}
\expandafter\def\csname rd@D2/inkling/pushback/GtoG\endcsname{44}
\expandafter\def\csname rd@D2/inkling/pushback/n\endcsname{52}
\expandafter\def\csname rd@D2/qwen/peer_sa/GtoF\endcsname{1}
\expandafter\def\csname rd@D2/qwen/peer_sa/n\endcsname{87}
\expandafter\def\csname rd@D2/qwen/pushback/GtoF\endcsname{46}
\expandafter\def\csname rd@D2/qwen/pushback/n\endcsname{55}
\expandafter\def\csname rd@E/conclusion/max\endcsname{6.4}
\expandafter\def\csname rd@E/member/option/max\endcsname{12.4}
\expandafter\def\csname rd@E/unchanged/max\endcsname{96.6}
\expandafter\def\csname rd@E/unchanged/min\endcsname{92.2}
\expandafter\def\csname rd@F/all/forced/challenger/ldtosd\endcsname{57.3}
\expandafter\def\csname rd@F/all/forced/challenger/sd\endcsname{26.7}
\expandafter\def\csname rd@F/all/forced/conclusion\endcsname{6.7}
\expandafter\def\csname rd@F/all/forced/inkling/d1_pp\endcsname{$-$3.9}
\expandafter\def\csname rd@F/all/forced/option\endcsname{7.7}
\expandafter\def\csname rd@F/all/forced/qwen/d1_pp\endcsname{$-$2.5}
\expandafter\def\csname rd@F/all/forced/qwen/d2_argument\endcsname{5.54}
\expandafter\def\csname rd@F/all/forced/qwen/d2_filler\endcsname{6.76}
\expandafter\def\csname rd@F/all/forced/recipient/ld\endcsname{76.0}
\expandafter\def\csname rd@F/all/forced/recipient/sa\endcsname{6.3}
\expandafter\def\csname rd@F/all/natural/challenger/sa\endcsname{52.3}
\expandafter\def\csname rd@F/all/natural/conclusion\endcsname{5.0}
\expandafter\def\csname rd@F/all/natural/inkling/d1_pp\endcsname{+0.9}
\expandafter\def\csname rd@F/all/natural/option\endcsname{6.4}
\expandafter\def\csname rd@F/all/natural/qwen/d1_pp\endcsname{+4.8}
\expandafter\def\csname rd@F/all/natural/qwen/d2_argument\endcsname{6.26}
\expandafter\def\csname rd@F/all/natural/qwen/d2_filler\endcsname{6.76}
\expandafter\def\csname rd@F/all/natural/recipient/ld\endcsname{25.7}
\expandafter\def\csname rd@F/all/natural/recipient/sa\endcsname{50.0}
\expandafter\def\csname rd@F/all/qwen/d1n\endcsname{50}
\expandafter\def\csname rd@F/all/qwen/d2n\endcsname{100}
\expandafter\def\csname rd@G/ce/concl\endcsname{10}
\expandafter\def\csname rd@G/ce/kappa\endcsname{0.87}
\expandafter\def\csname rd@G/ce/n\endcsname{50}
\expandafter\def\csname rd@G/ce/same\endcsname{48}
\expandafter\def\csname rd@G/human/kappa\endcsname{0.80}
\expandafter\def\csname rd@G/human/n\endcsname{50}
\expandafter\def\csname rd@G/human/same\endcsname{43}
\expandafter\def\csname rd@G/opus/kappa\endcsname{0.83}
\expandafter\def\csname rd@G/opus/n\endcsname{50}
\expandafter\def\csname rd@G/opus/same\endcsname{44}
\expandafter\def\csname rd@paths/dyad\endcsname{300}
\expandafter\def\csname rd@paths/triad\endcsname{600}
\expandafter\def\csname rd@questions\endcsname{50}

\newcommand{\layerA}{\textsc{Layer\,A}}\newcommand{\layerB}{\textsc{Layer\,B}}\newcommand{\layerC}{\textsc{Layer\,C}}\newcommand{\layerD}{\textsc{Layer\,D}}\newcommand{\layerE}{\textsc{Layer\,E}}

\title{What Does Multi-Agent Debate Actually Change?}

\author{Chen Qian\\\texttt{chenqian@andrew.cmu.edu}}
\date{}

\begin{document}
\maketitle

\begin{abstract}
Multi-agent debate, in which several LLMs exchange arguments before producing
an answer, raises a basic question: does expressed disagreement reflect changes
in the members' own positions? No single signal can settle this question,
so we organize the analysis around five questions:
(A)~does the debater \emph{say} it disagrees;
(B)~does its reply text actually \emph{argue};
(C)~does its own position \emph{change} after each debate turn;
(D)~how much, quantitatively, does the position change after debate turns; and
(E)~how do members' \emph{final positions} compare with their initial ones?
We evaluate two- and three-member committees on 50 curated opinion questions
from GlobalOpinionQA, using same-model, same-family, and mixed-family
configurations with \emph{friendly} (seek common ground), \emph{neutral},
and \emph{hostile} (stress-test positions) instructions assigned at each turn.
The answers differ.
(A)~Tone changes reported agreement: across configurations, the share of replies
reporting strong agreement is 70.7--96.3\% under friendly instructions,
compared with 9.5--19.3\% under hostile ones, with model choice shaping
the size of the difference.
(B)~Self-reports broadly align with text judgments, but consistency varies
by agreement level and model choice. For example, in two-member debates,
a text evaluator rates about one in four replies labeled \emph{leaning agree}
as \emph{strongly agree}.
(C)~Position changes depend on the interaction and model choice. For example,
in two-member debates, after a peer's leaning-disagree reply, members reporting
strong agreement switch options more often than those reporting
leaning disagreement.
(D)~For open-weight members in two-member mixed-family committees, the
probability of the option a member already holds stays near saturation and
barely changes within a turn, even after a peer's pushback. In a separate probe, when consecutive replies both
report leaning disagreement, endorsement of earlier position text is lower
after a peer's argument than after neutral filler, with a larger decrease
for Qwen3.8-27B than Inkling. These measure contextual responses, not lasting
beliefs.
(E)~Comparing initial positions with those at each member's last participation
on a debate path, we find that the selected option is unchanged in over 90\%
of comparisons in every configuration.
Taken together, during multi-agent debate, expressed disagreement need not
translate into position revision, either within individual exchanges or
over a complete debate.
\end{abstract}

\section{Introduction}
Multi-agent LLM debate brings several models together to exchange arguments before producing a final
answer~\citep{du2023debate,liang2023mad}. The motivation is that models can challenge one another's
reasoning and use the discussion to improve their answers. However, models may also agree through
sycophancy or conformity~\citep{sharma2023sycophancy,perez2022discovering,zhu2024conformity}, abandon
a stated position under pushback that does not justify the change~\citep{wang2023defend,xu2024earthflat},
and weaker models may yield to more assertive peers~\citep{xiong2023ford}; debates among similar models
can also settle on a shared majority view~\citep{estornell2024multillm}. On tasks with known answers,
debate does not consistently outperform a single well-prompted model or a matched sample-and-vote
baseline~\citep{smit2023mad,li2024moreagents,wang2024rethinking,kenton2024debate}.
These findings raise a broader question: what does the exchange of arguments actually change?

On verifiable tasks, a flawed argument can be checked against the answer, so a discussion can quickly
converge on whichever member is right, or on whichever member sounds most
confident~\citep{xiong2023ford,kenton2024debate}. Open-ended questions remove this anchor: with no
reference answer to converge to, any agreement or disagreement has to arise from the exchange itself.
We therefore focus on \emph{open-ended} opinion questions and study how debate produces disagreement and
whether it changes the members' positions. We investigate these questions through a layered analysis of
reported agreement, the reply text, turn-level position changes, token-probability responses, and the
change from each member's initial position to its final one.

\paragraph{Measuring disagreement and position change.}
We use GlobalOpinionQA~\citep{durmus2023globalopinions}, which asks for views on political and social
issues, each with a predefined list of answer options. A committee of two or three LLM members first
forms independent opinions and then responds to one another's arguments. A router decides which member
replies to which earlier contribution, and we call each resulting chain of replies a \emph{debate path}.
Within a path, each \emph{debate turn} is a reply to one specific earlier contribution from another
member, with the earlier turns of the path provided as context.
Alongside its reply, the member reports how much it agrees with the contribution it is answering, and,
in the same generation, its current answer option and position on the original question. For each debate
turn, we assign a friendly, neutral, or hostile tone instruction (Section~\ref{sec:design}), so that we can
measure how instructions shape the debate. Each debate runs for a fixed number of turns.

\begin{figure}[t]
\centering
\includegraphics[width=\textwidth]{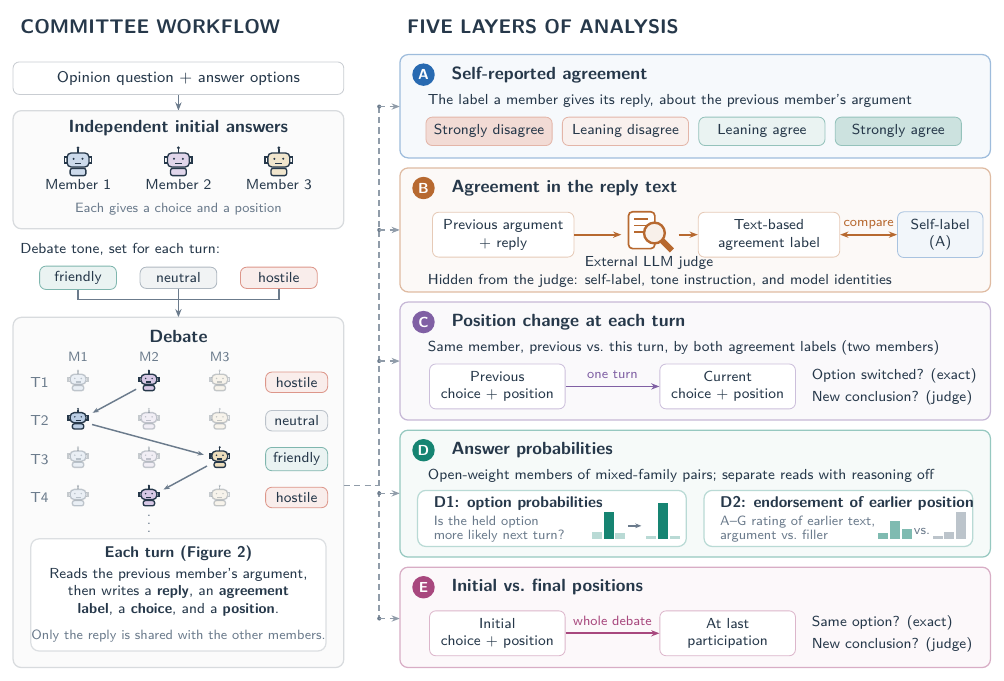}
\caption{\textbf{Committee workflow and the five layers of analysis.}
Two- or three-member committees (same-model, same-family, or mixed-family) answer independently, then debate
with a friendly, neutral, or hostile tone set for each turn. Layers A--E are parallel readings of the same
debates; \layerC{} uses two-member committees, and \layerD{} the open-weight members of mixed-family pairs.
In the debate panel, each row is one turn: the highlighted member speaks, replying to the previous turn's speaker (two-member debates alternate between two members). Speakers, tones, and probability bars are illustrative.}
\label{fig:framework}
\end{figure}

\begin{figure}[t]
\centering
\includegraphics[width=0.62\textwidth]{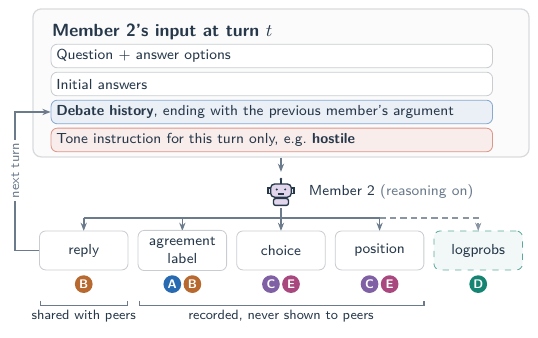}
\caption{\textbf{One debate turn.}
A member sees the question, the initial answers, and the debate history ending with the previous member's argument, and
this turn's tone, then writes a reply, an agreement label, a choice, and a position in one generation. Only the
reply enters later turns. Badges mark which layer uses each output; \layerD{} reads token probabilities
(logprobs) for the same input in a separate request.}
\label{fig:turn}
\end{figure}

We formalize the measurement through five layers of analysis (Figures~\ref{fig:framework}
and~\ref{fig:turn}):
\begin{itemize}
  \item \layerA{}: the agreement level the member reports for its own reply, one of \emph{strongly agree},
        \emph{leaning agree}, \emph{leaning disagree}, and \emph{strongly disagree}.
  \item \layerB{}: whether the reply \emph{text} expresses that level of agreement, judged on the same scale by
        an external grader that sees neither the self-reported label nor the tone instruction.
  \item \layerC{}: whether each member's position changes at a debate turn. For example, if a member reports
        leaning agreement with a peer's pushback against its previous position, does its own answer change?
  \item \layerD{}: how much, quantitatively, each member's position moves during the debate, measured
        through the probabilities the model assigns to the answer options and to its earlier position.
  \item \layerE{}: whether the debate as a whole changes positions, comparing each member's initial position
        (zero debate turns) with its position at its last turn on a debate path.
\end{itemize}

To understand the effect of model choice, we use three committee configurations:
\begin{itemize}
  \item \textbf{Same model:} all members use the same model.
  \item \textbf{Same family:} members use different models from one family (GPT-5.6 Luna, Terra, and Sol).
  \item \textbf{Mixed family:} members use models from different families (GPT-5.6 Terra, Qwen3.8-27B, and Inkling).
\end{itemize}
In each configuration, we also run exclusive two-member debates. These support \layerC{}: in a two-member
path, the incoming contribution always comes from the one other member and directly answers the receiver's
own previous turn. For the quantitative measurement in \layerD{}, we use the open-weight members of the
mixed-family committee, whose output-token probabilities we can read.

\paragraph{Contributions.}
We study two- and three-member committees in three configurations on \rd{questions} open-ended opinion
questions from GlobalOpinionQA~\citep{durmus2023globalopinions}.
Our committee implementation and all five evaluation layers are publicly
available.\footnote{\url{https://github.com/chenmoneygithub/llm-committee}}
Our main contributions are:
  \begin{enumerate}
  \item \textbf{\layerA{}: self-reported agreement.}
  We record how much each member reports agreeing with the argument it is answering, and compare these
  labels across friendly, neutral, and hostile turns. Tone strongly changes the reported agreement level:
  the share of replies reporting strong agreement is \rd{A/friendly/sa/min}--\rd{A/friendly/sa/max}\%
  under friendly instructions and \rd{A/hostile/sa/min}--\rd{A/hostile/sa/max}\% under hostile ones.
  Model choice shapes what replaces strong agreement: under hostile instructions, same-model members mostly
  move to \emph{leaning agree} (\rd{A/dyad/same/hostile/la}\% in two-member debates), whereas mixed-family
  members report \emph{leaning disagree} far more often (\rd{A/dyad/mixed/hostile/ld}\%, versus
  \rd{A/dyad/same/hostile/ld}\% for the same model).

  \item \textbf{\layerB{}: disagreement in the reply text.}
  We assess whether reported agreement also appears in what the members write.
  A judge from outside the committee's model families reads each reply together with the argument it
  answers, without seeing the structured self-reported label or the tone instruction.
  Self-reports and text judgments align closely in every configuration (exact agreement
  \rd{B/exact/min}--\rd{B/exact/max}\%, Cohen's $\kappa$ \rd{B/kappa/min}--\rd{B/kappa/max}), but not
  perfectly, and the alignment varies by agreement level and model: the judge rates
  \rd{B/latosa/min}--\rd{B/latosa/max}\% of replies self-labeled \emph{leaning agree} as \emph{strongly
  agree}, and alignment is lowest in mixed-family committees.

  \item \textbf{\layerC{}: position changes.}
  We measure how often members change their answer option or position text after a debate turn, grouped
  by the peer's and their own agreement labels. Positions change far less than reported agreement:
  across all sampled two-member turns, members switch options in \rd{C/same/option/all}\% (same model),
  \rd{C/family/option/all}\% (same family), and \rd{C/mixed/option/all}\% (mixed family) of comparisons.
  Switches concentrate in one interaction: after a peer's \emph{leaning disagree} reply, members that
  report \emph{strongly agree} switch options in \rd{C/same/ld/sa/option}\%, \rd{C/family/ld/sa/option}\%,
  and \rd{C/mixed/ld/sa/option}\% of cases, versus at most \rd{C/ld/ld/option/max}\% for members that
  report \emph{leaning disagree}
  (Section~\ref{sec:results}).

  \item \textbf{\layerD{}: position strength changes.}
  As an extension of \layerC{}, we read the probabilities of the predefined GlobalOpinionQA options before
  each debate turn, and separately the probability-weighted endorsement of the member's earlier position
  text after a peer's argument versus neutral filler (Section~\ref{sec:logprob}).
  Once a member has stated its answer, the probability of that option stays near saturation and barely
  moves within a turn, even after pushback (\rd{D1/qwen/later_pushback/change}pp for Qwen3.8-27B and
  \rd{D1/inkling/later_pushback/change}pp for Inkling). Endorsement of the earlier position text does respond:
  after pushback, Qwen3.8-27B's most likely rating of its earlier position drops from \emph{completely agree} to
  \emph{mostly agree} in \rd{D2/qwen/pushback/GtoF} of \rd{D2/qwen/pushback/n} cases, while Inkling's rarely
  moves. The conclusion holds; the endorsement of its reasoning softens.

  \item \textbf{\layerE{}: changes of the final position.}
  We compare each member's initial position with its position at its last participation on a debate path.
  The selected option is unchanged in \rd{E/unchanged/min}--\rd{E/unchanged/max}\% of comparisons across
  configurations.
  \end{enumerate}
Complementing \citet{hao2026flips}, who separate spontaneous answer changes from peer influence, we separate
expressed agreement from position change in multi-LLM debate; we do not claim to be the first to quantify
committees. We progressively separate
reported, textual, turn-level, probability-level, and end-of-debate responses, and use a judge from outside
the committee's model families, compared against human annotations. We build on
LMC~\citep{zhao2024lmc}, ChatEval~\citep{chan2023chateval}, Peacemaker~\citep{peacemaker2025}, and
Self-MoA~\citep{li2025selfmoa} (\S\ref{sec:related}).

\paragraph{The takeaway, and what this paper does \emph{not} claim.} In plain terms, the findings
are three. First, debate tone strongly changes how much agreement the committee \emph{reports}, and the
reply text largely corroborates these reports, with model choice shaping both. Second, members' positions
change much less than their reported agreement, both in the turn-level statistics and in the open-weight
members' probabilities. Third, over a complete debate, members' final answers rarely differ from their
initial ones: expressed disagreement need not translate into position revision.

We do not evaluate whether debate improves final-answer quality; on open-ended questions we focus on whether
members' opinions change.

\section{Related Work}
\label{sec:related}
\paragraph{Debate and final-answer quality.}
Multi-agent debate serves two related but distinct purposes.
In debate for oversight, a judge uses competing arguments to assess answers it may be unable to
produce or verify unaided~\citep{irving2018debate,michael2023debate,khan2024persuasive}.
In debate for answer generation, models exchange and revise responses to improve their joint
output~\citep{du2023debate,liang2023mad}.
These studies cover more than exact-answer tasks: MAD also evaluates machine translation using
automatic metrics and human ratings~\citep{liang2023mad}.
Here, \emph{open-ended opinion questions} means questions without a unique correct answer, not
simply tasks that require free-form text.
Our study uses such questions to examine disagreement and position change separately from output quality.

Evidence that debate improves answers depends on the task, protocol, and baseline.
Several evaluations find that debate does not reliably outperform a well-prompted single model
or repeated independent sampling~\citep{smit2023mad,li2024moreagents,wang2024rethinking,zhang2025overvaluing}.
In an oversight setting, \citet{kenton2024debate} find that debate outperforms direct question
answering on extractive tasks where debaters see a source article that the judge cannot access;
results on other tasks without this information asymmetry are mixed.
These results concern the accuracy of an outcome; we instead ask what the exchange changes in the
participants, which can be measured even when no answer is correct.

The choice of baseline also determines what an improvement can establish.
Mixture-of-Agents combines outputs through successive aggregation
stages~\citep{wang2024moa}, whereas Self-MoA shows that aggregating repeated outputs from a strong
single model can outperform mixing models~\citep{li2025selfmoa}.
Equal-budget comparisons can favor debate or aggregation at some budgets and self-consistency at
others~\citep{wunderlich2026pareto}. Related studies examine limits imposed by shared model
errors~\citep{chen2026cofailure} and cases where a team underperforms its expert
member~\citep{pappu2026teams}; open-source council experiments also compare governance structures
with self-consistency and prompt-diversity baselines~\citep{andybhall2025council}.
These results show that model composition matters separately from interaction.
We do not compare debate with aggregation baselines; instead, we hold the protocol fixed and vary
composition, from three instances of one model to models from different families, to ask how model
similarity shapes the disagreement members express and the positions they hold.

Interaction schedules provide another source of variation.
Repeated peer revision appears in early debate protocols~\citep{du2023debate,liang2023mad};
layered aggregation provides a related, but not identical, form of cross-model
interaction~\citep{wang2024moa} (see also \citealp{tran2025survey}).
Longer discussions need not be better: \citet{kenton2024debate} find no significant benefit from
additional turns in their tested protocols, \citet{becker2025drift} examine how discussion
can drift away from the task, and debates among similar models can converge on a shared majority
view~\citep{estornell2024multillm}.
Our two-member paths alternate between the same pair of members, whereas our three-member paths
are routed among all members; in both, we record every member's position at every turn rather than
only the final output.

\begin{table}[t]
\centering\small
\setlength{\tabcolsep}{4pt}
\renewcommand{\arraystretch}{1.12}
\begin{tabular}{@{}p{0.21\textwidth}p{0.34\textwidth}p{0.40\textwidth}@{}}
\toprule
\raggedright \textbf{Study} & \raggedright \textbf{Main measurement} & \raggedright \textbf{Connection to this study} \tabularnewline
\midrule
\raggedright \citet{du2023debate}; MAD~\citep{liang2023mad}
& \raggedright Answer quality after exchanging and revising responses.
& \raggedright We measure disagreement and position change on opinion questions, not answer quality. \tabularnewline
\addlinespace[3pt]
\raggedright \citet{smit2023mad}; \citet{zhang2025overvaluing}
& \raggedright Debate versus single-model and independent-sampling baselines.
& \raggedright Motivate testing, not assuming, the value of interaction. \tabularnewline
\addlinespace[3pt]
\raggedright \citet{kenton2024debate}
& \raggedright Weaker judges using stronger debaters, with or without private evidence.
& \raggedright Oversight differs from peer debate on questions without a correct answer. \tabularnewline
\addlinespace[3pt]
\raggedright MoA; Self-MoA~\citep{wang2024moa,li2025selfmoa}
& \raggedright Quality of aggregated outputs, including same-model resampling.
& \raggedright Motivate varying model composition separately from interaction. \tabularnewline
\addlinespace[3pt]
\raggedright LMC; PoLL~\citep{zhao2024lmc,verga2024poll}
& \raggedright Agreement between model juries and human ratings.
& \raggedright Inform judge design; our human check concerns agreement in replies. \tabularnewline
\addlinespace[3pt]
\raggedright ChatEval~\citep{chan2023chateval}
& \raggedright Human agreement when evaluators debate an external answer.
& \raggedright Their evaluators debate; ours rates disagreement in the debate under study. \tabularnewline
\addlinespace[3pt]
\raggedright Peacemaker~\citep{peacemaker2025}
& \raggedright Disagreement collapse, correct-position abandonment, and sycophancy.
& \raggedright We measure whether positions change, not whether a change is correct. \tabularnewline
\addlinespace[3pt]
\raggedright \citet{hao2026flips}
& \raggedright Answer changes after self-reflection, peer stances, or peer reasoning.
& \raggedright Both separate peer influence from re-answering; we compare arguments with filler. \tabularnewline
\addlinespace[3pt]
\raggedright \citet{keramati2026liar}
& \raggedright Reasoning-token probabilities, judge ratings, and accuracy.
& \raggedright We read option and endorsement probabilities of the debating members. \tabularnewline
\midrule
\raggedright This study
& \raggedright Reported and textual agreement; turn-level and end-of-debate position change; option and endorsement probabilities.
& \raggedright All five layers read the same debates; answer quality is not evaluated. \tabularnewline
\bottomrule
\end{tabular}

\caption{Selected studies, their primary measurements, and their relationship to our five-layer analysis.
Opinion questions without a unique correct answer are distinct from free-form generation tasks
such as translation.}
\label{tab:related}
\end{table}

\paragraph{Disagreement, conformity, and position change.}
Sycophancy studies show that models can align responses with a user's stated
views~\citep{sharma2023sycophancy,perez2022discovering}.
Related work examines conformity under social influence~\citep{zhu2024conformity}, the role of
assertiveness in model-to-model debate~\citep{xiong2023ford}, and the propagation of sycophancy or
loss of expressed disagreement across agents and rounds~\citep{kasprova2026polite,wynn2025talk}.
Models also abandon correct answers when a debate partner pushes back with invalid
arguments~\citep{wang2023defend}, and multi-turn persuasion can shift their stated beliefs even about
facts they know~\citep{xu2024earthflat}.
Studies of user pushback and rebuttal timing likewise show why a changed answer need not reflect
improved reasoning~\citep{muse2026,fanous2025syceval}.
At the population level, networks of LLM agents tend to converge in opinion
simulations~\citep{chuang2024opinion} and can develop shared conventions and collective biases absent
in any single agent~\citep{flintashery2025conventions}.
These findings motivate our comparison of friendly, neutral, and hostile instructions. Role and persona
instructions have effects that are hard to predict~\citep{zheng2024personas}, so we measure the effect
of tone rather than assuming it.

Peacemaker or Troublemaker~\citep{peacemaker2025} connects debate dynamics to correctness and
sycophancy under different personas.
Its Negative Agreement Rate measures abandonment of a correct position during disagreement.
On opinion questions, remaining consistent is not necessarily correct, and changing
position is not necessarily an error; we therefore measure whether positions change, not whether
a change is right.

A closely related measurement problem is that an answer can change simply because the model is
asked again. \citet{hao2026flips} compare self-reflection, exposure to peers' stances alone, and
exposure to their full reasoning to distinguish spontaneous answer changes from peer influence.
We address the same problem in two ways. Our \layerC{} records the choice and position a member emits in
the same generation as its reply, rather than re-asking afterward, and our \layerD{} endorsement probe
compares a peer's actual argument with length-matched filler in otherwise identical inputs.

Work on instruction stability and repeated questioning provides further reasons to test whether
a stated position persists~\citep{li2024instability,xie2023askagain}.
Faithfulness studies show that generated explanations need not reveal what influenced a
response~\citep{turpin2023languagemodels,lanham2023faithfulness}, and models have only limited
introspective access to their own behavior~\citep{binder2025introspection}.
We draw a narrower distinction: the model's agreement label, the agreement expressed in its reply,
and changes in its own answer are separately measured outcomes.
None alone establishes an underlying belief or the reasoning process that produced the answer.

\paragraph{Evaluating replies.}
A model committee can also act as an evaluator rather than as the system being evaluated.
Language Model Council compares aggregated model judgments with human preferences on subjective
tasks~\citep{zhao2024lmc}; PoLL combines judges from different model
families~\citep{verga2024poll}.
ChatEval uses discussion among evaluators to assess an external
answer~\citep{chan2023chateval}, while DEBATE uses a devil's-advocate evaluator to challenge an
initial assessment~\citep{kim2024debate}.
These methods inform our use of model judges, but evaluating a completed answer and measuring
disagreement within the discussion are different tasks.

We use one judge, from outside the committee's model families, to rate agreement in replies and to
compare position texts. For agreement, it sees the reply and the contribution it answers, but not the
self-reported label or tone instruction.
We compare its ratings with human annotations. This is a task-specific comparison, not evidence
that model judges can generally replace people; JUDGE-BENCH documents substantial variation across
evaluation tasks~\citep{bavaresco2024judgebench}.

\paragraph{Token probabilities and stance options.}
Token probabilities over fixed answer options have been used to measure model
opinions~\citep{santurkar2023opinions} and confidence~\citep{kadavath2022know}, although verbalized
confidence and token probabilities can disagree~\citep{tian2023justask,xiong2024uncertainty}.
\citet{keramati2026liar} study the relationship between log probabilities of generated reasoning
tokens, external judgments of that reasoning, and final-answer accuracy.
Our \layerD{} instead reads the probabilities that debate members themselves assign to the original
answer options and to a seven-point endorsement scale for their earlier position.
Comparisons with approximately length-matched filler help distinguish responses to an argument's
content from the effects of adding text.
These are contextual response probabilities, not calibrated confidence or direct measurements of beliefs
(Section~\ref{sec:logprob}).

\paragraph{Opinion dynamics and belief change in multi-agent systems.}
Classical opinion-dynamics models update each agent's opinion as a weighted average of others'
opinions~\citep{degroot1974consensus}, or only of opinions within a confidence
bound~\citep{deffuant2000mixing,hegselmann2002opinion}; in these models, the opinion an agent communicates and the one
it holds are the same quantity. Formal persuasion dialogues track the public commitments agents incur through their
moves~\citep{prakken2006persuasion}, and belief revision specifies how a rational agent should change its beliefs given
new information~\citep{alchourron1985agm}. LLM agents give no direct access to such an internal state, so whether a
stated agreement reflects an updated position has to be measured; our layers separate what an agent says (\layerA{},
\layerB{}) from the answer it holds (\layerC{}, \layerE{}) and the probabilities behind that answer (\layerD{}).

Together, these literatures motivate our separation of reported agreement, agreement expressed in
text, turn-level position change, option and endorsement probabilities, and end-of-debate position
change (Table~\ref{tab:related}).

\section{Experimental Setup}
\label{sec:data}
\subsection{Data}
Our experiments use opinion questions from GlobalOpinionQA~\citep{durmus2023globalopinions}, which
collects multiple-choice questions from cross-national surveys.
These questions have no unique correct answer. We therefore evaluate how models disagree and how their
positions change, without treating any survey population's answer as ground truth.

Many of the dataset's questions do not apply naturally to an LLM committee. Some ask respondents to report
personal experiences, habits, or identities rather than opinions; others presuppose a national context
that the question does not state, e.g., ``Does this statement describe your country very well, somewhat
well, not too well, or not well at all? Elected officials care what ordinary people think.''
We therefore curated a bank of 60 questions that can actually be debated (Appendix~\ref{app:questions}).
We retain questions that ask for opinions rather than reports of respondents' personal experiences, habits,
or identities. For example, we include a question asking whether respondents agree with the statement
``A university education is more important for a boy than for a girl,'' but exclude the personal-history
question ``In the last five years have you traveled to another country other than Canada or Mexico?''
Three LLM classifiers (GPT-5.5, Claude Opus 4.8, and Gemini 3.5 Flash) labeled each candidate in the
dataset's order, and we kept candidates that at least two labeled as opinion questions; we then removed
questions whose answer depends on an unspecified country or government.
Question wording and answer options are never edited.
The main study uses \rd{questions} of the 60 questions, sampled uniformly at random before any debate was
run; no selection step uses debate outcomes.

\label{sec:method}
\subsection{Tasks and committee workflow}
\label{sec:framework}

We use three committee configurations to study the effect of model choice on the debate:
\begin{itemize}
\item \textbf{Same model:} three members that all use GPT-5.6 Terra, treated as distinct members with
      separate generations and histories.
\item \textbf{Same family:} GPT-5.6 Luna, GPT-5.6 Terra, and GPT-5.6 Sol, three models from one family.
\item \textbf{Mixed family:} GPT-5.6 Terra, Qwen3.8-27B, and Inkling, three models from different developers.
\end{itemize}
GPT-5.6 members use medium reasoning, and the open-weight members also reason before answering in every
debate turn. Apart from the roster, all configurations share the same questions, prompts, routing, and tone
schedules.

Each committee is run in a two-member mode and a three-member mode, which support different layers.
Before any debate, every member independently answers the question with a \emph{choice} (one of the original
answer options, or none if no option fits) and a \emph{position}, a paragraph stating its view and reasons.
A router then sends one member's contribution to another member, and each chain of replies forms a
\emph{debate path}.
In the two-member mode, each of the three member pairs debates separately: the members alternate for four
turns, starting with a reply to the other member's initial answer. A question therefore has three
two-member paths per tone assignment.
In the three-member mode, a fixed routing tree per question, drawn at random before generation, starts
from two different initial exchanges and branches into six paths of five turns each; paths share the
replies on their common prefix.

On each \emph{debate turn}, a member replies to one specific contribution from a peer
(Figure~\ref{fig:turn}). Its input contains the question and options, the initial answers of the members on
the path, the replies earlier on the path, the incoming contribution, and the current turn's tone
instruction. In one generation, the member returns four fields: a \emph{reply} to the incoming contribution,
an \emph{agreement} label for that contribution, and its current \emph{choice} and \emph{position} on the
original question. Only the reply is added to later turns' history. Agreement labels, choices,
positions, tone instructions, and all measurements are recorded for analysis but never shown to other
members, so later turns are not steered by the recorded state.

For example, in a mixed-family debate on the university-education question, Inkling and GPT-5.6 Terra
both initially choose \emph{completely disagree}. Under a hostile instruction, Terra endorses Inkling's
conclusion but challenges two of its reasons, that university suits everyone and that it is a fundamental
right, and labels its reply \emph{strongly agree}. Inkling, also under a hostile instruction, accepts both
corrections, labels its reply \emph{leaning agree}, keeps its choice, and revises its position text to
concede the rights framing. The label concerns agreement with the peer's message, not agreement with the
question's claim, and a member can revise its reasons without changing its answer.
Appendix~\ref{app:transcript} shows this exchange.

Our implementation can also pass a debate to a \emph{chairman} model that synthesizes the members' views into
one final answer. Because this paper studies the members' own positions rather than final-answer quality,
no layer uses the chairman's output.
Appendix~\ref{app:impl} gives the routing and implementation details.
The separate protocol needed to read token probabilities is described in
Section~\ref{sec:method-logprob}.

\paragraph{Tone assignment.}
\label{sec:design}
For each debate turn, we assign one of three tone instructions to study how tone changes the debate:
\begin{itemize}
\item \textbf{Friendly}: seek common ground and concede when the other contribution has merit.
\item \textbf{Neutral}: use the base debate prompt without an additional tone instruction.
\item \textbf{Hostile}: stress-test each position and avoid agreement without a substantive reason.
\end{itemize}
Tones are assigned to turns before generation, balanced across questions within each pair (or routing
position) and turn index, and independent of the preceding messages.
A member sees only the instruction for its current turn; earlier turns' instructions are not shown, although
the earlier replies may reflect their tones. Initial answers use no tone instruction, so the manipulation
concerns engagement with peers, not the answer options.
The exact tone additions appear in Appendix~\ref{app:prompts}.

For each path, we generate two tone assignments that share the first two turns and differ at every later
turn. For example, a two-member path whose shared turns T1 and T2 are hostile may continue with hostile and
friendly turns in one assignment and with neutral and hostile turns in the other.
This doubles the number of debates per question while holding the opening of the exchange fixed.
Each configuration thus yields \rd{paths/dyad} two-member paths with \rd{A/dyad/replies} unique replies and
\rd{paths/triad} three-member paths with \rd{A/triad/replies} unique replies.
Paths and tone assignments are not independent samples: questions are the unit of independent sampling.

\subsection{Reported and textual agreement: Layers A and B}
We first report statistics of the self-reported agreement level under different tones and configurations,
along with how well the reply text aligns with the self-reported level.

\paragraph{Self-reported agreement (A).}
Each reply includes an agreement label with four levels: \emph{strongly agree}, \emph{leaning agree},
\emph{leaning disagree}, and \emph{strongly disagree}. A shared rubric defines each level relative to the
peer's central position: \emph{strongly} denotes substantive overall endorsement or rejection, regardless of
politeness or rhetorical force, and \emph{leaning} denotes support or opposition that is materially qualified
(Appendix~\ref{app:prompts}).
We compare these label distributions across tones and configurations, focusing on the share of replies
labeled strongly agree. Members may return a null label if they cannot determine an overall direction; only
explicitly emitted labels enter the analysis. A analyzes every reply, including those at T1.

\paragraph{Agreement expressed in text (B).}
To assess the reply separately from its label, a judge from a different model family than every committee
member reads the question, the incoming contribution, and the reply, without seeing the member identity,
tone instruction, or structured self-report. Self-descriptions embedded in the reply itself are not removed
(Appendix~\ref{app:transcript}). It rates agreement with the same four-level rubric that the members use, or
marks the reply as having no position or as unjudgeable.

The judge is Gemini 3.8 Flash. The committee members are OpenAI's GPT-5.6 models, Alibaba's Qwen3.8-27B, and
Thinking Machines' Inkling, so the judge shares a model family with none of them.
To keep judging costs bounded, B judges eight turns per question, sampled before generation (T2--T4 for two
members and T2--T5 for three) and judged in both tone assignments, with shared turns judged once:
\rd{B/dyad/same/n} replies per two-member configuration and \rd{B/triad/same/n} per three-member
configuration. We report the full self-label by judge-label table, its exact agreement, and Cohen's $\kappa$.
The rubric and judge prompt are in Appendix~\ref{app:judge}.

\paragraph{Grounding the model judge.}\label{sec:grounding}
We ground the judge on 50 sampled replies, stratified by self-label and spread across all six settings
(Appendix~\ref{app:judge}). A second LLM judge (Claude Opus 5.5), given identical inputs, agrees with it on
\rd{G/opus/same} of \rd{G/opus/n} ($\kappa={}$\rd{G/opus/kappa}), and one of the authors verified the judge's labels,
agreeing with \rd{G/human/same} of \rd{G/human/n} ($\kappa={}$\rd{G/human/kappa}). For the position-text judge used in
C and E, the second judge agrees on whether the conclusion changed in \rd{G/ce/same} of \rd{G/ce/n} sampled
comparisons ($\kappa={}$\rd{G/ce/kappa}).

\subsection{Measuring position changes: Layers C and D}
Layers A and B compare reported and textual agreement across debate turns conducted under different tone
instructions. Layers C and D study position shifts: C reports how often a member's choice or position
changes, and D quantifies the movement through the member's answer probabilities.

To make the source of each argument unambiguous, C and D use the two-member paths of each configuration.
In a two-member path, the incoming contribution always comes from the one other member and responds directly
to the receiver's own previous turn, so we can relate a member's position change to the argument it received.

\paragraph{Position change (C).}
After each debate turn, we compare the member's choice and position with those it reported at its previous
participation on the same path, or with its initial answer at its first participation. A change in the
predefined choice is computed directly. For the position text, the external judge sees only the question,
options, and the two positions, and rates whether the main conclusion changed.
C uses the same sampled turns as B (T2--T4).

For reporting, we break down each turn by the agreement label of the peer's preceding reply and the member's
own label in the current turn, up to $4\times4=16$ combinations. These labels are observed outcomes rather
than randomized conditions, so the breakdown describes association, not the causal effect of a label.

\paragraph{Position strength change (D).}
A member's position can be weakened or reinforced through debate without changing its choice, which C's
yes/no comparison cannot capture. We therefore read output probabilities from the open-weight members to
measure the strength of their positions with two probes:
\begin{enumerate}
\item \textbf{D1, option probabilities}: how the member's probabilities over the original answer options change
      between consecutive turns; and
\item \textbf{D2, endorsement of the earlier position}: how much the member endorses its own earlier position text
      after a peer's argument, relative to neutral filler.
\end{enumerate}
D1 reports the probability change of the earlier choice; D2 reports the change in the expected rating.

\label{sec:method-logprob}
\paragraph{Probability reads.}
Following work that reads model opinions and confidence from the probabilities of fixed answer
options~\citep{santurkar2023opinions,kadavath2022know}, we measure \layerD{} on the open-weight members of the
mixed-family two-member committees, Qwen3.8-27B and Inkling, whose serving endpoint returns token
log probabilities.
The reads are separate requests that never enter the debate. Each reuses the input of the corresponding
debate turn, but replaces the tone instruction and output format with a neutral request for a single
letter, and disables reasoning so that the letter is the first generated token.
We take the returned top-20 log probabilities at that position, assign zero probability to any candidate
letter outside the top 20, and renormalize over the candidate letters.
These are probabilities of a model's next answer in a given context, not measurements of an internal
belief~\citep{tian2023justask}.

\paragraph{Option probability (D1).}
Before each sampled turn, the member reads the same discussion it is about to answer and selects one of
the original GlobalOpinionQA options. It does not see the choice, reply, or position it generates in that
turn. Let $c$ be the option the member selected at its previous participation $t'$ on the path, and let $p_t$
be the option distribution read before turn $t$.
We report the change in probability of the previously held option,
\begin{equation}
\Delta p_{\mathrm{D1}} = p_t(c) - p_{t'}(c),
\label{eq:d1}
\end{equation}
in percentage points; a positive value means more support for the option the member already held.
Unlike D2's rating scale, GlobalOpinionQA options are not always ordered (e.g., a choice among religions), so
we do not compute an expectation over them.
We compute D1 only from a member's second participation on, so that both reads already contain the member's own
earlier answer. At its first participation, the earlier read is the question alone, and the comparison would
mostly measure the effect of having stated an answer. Because the earlier and later inputs still differ by more
than the incoming argument, D1 describes how support evolves along the debate; it does not isolate the causal
effect of one argument.

\paragraph{Endorsement of the earlier position (D2).}
Additionally, we measure how much a member still endorses its own position just before a debate turn.
We give the member two inputs that are identical except for the incoming contribution: the \emph{argument}
input contains the peer's actual reply, and the \emph{filler} input replaces it with a neutral procedural
sentence repeated to approximately match its token length. Both inputs append the member's position text
from before the turn, and ask it to rate that fixed text with one letter from A (completely disagree) to G
(completely agree). Encoding A--G as $k=1,\ldots,7$, we compute the expected rating
$\mu(p)=\sum_{k=1}^{7}k\,p(k)$ under each input and report
\begin{equation}
\Delta_{\mathrm{D2}} = \mu\!\left(p^{\mathrm{argument}}\right) - \mu\!\left(p^{\mathrm{filler}}\right),
\label{eq:d2}
\end{equation}
in rating points; a negative value means the argument lowers endorsement of the member's earlier position.
A drop of one point corresponds, for example, to moving from \emph{completely agree} to \emph{mostly agree}.
For both D1 and D2, we average comparisons within each question and then weight questions equally, separately
for each model and each combination of agreement labels.
The prompts and examples of the full probability readouts are in Appendix~\ref{app:logprob}.

\subsection{How debate changes opinions: Layer E}
\label{sec:method-e}
Layer E asks how a complete debate changes positions, comparing each member's initial position (zero debate
turns) with its position at its last participation on each debate path. Last participation may precede the
path's final turn, and members that never reply on a path are excluded rather than counted as unchanged.
We directly compare the predefined choices for all available endpoints, and the external judge labels the
position texts as in C, on endpoints sampled before generation. E covers both two- and three-member committees in all configurations, and we
report it per member, without grouping by the final turn's tone or agreement labels.

\subsection{Supplementary study: forced disagreement}
\label{sec:method-forced}
In the main study, members choose how much to disagree. To test whether a member revises its position when a
peer substantively opposes it, we run a supplementary intervention on the mixed-family two-member committees.
Instead of regenerating the debate from scratch, we reuse the collected trajectories and intervene at selected turns. At a chosen turn, the replying member (the \emph{challenger}) receives a private instruction to take the opposing
side to the central conclusion of the message it is answering, while keeping that turn's scheduled tone
(Appendix~\ref{app:prompts}). The instruction does not prescribe the agreement label, and it is never shown
to the other member (the \emph{recipient}) or to the judge. The recipient then replies under its own scheduled
tone, with no additional instruction, and the debate stops.

\section{Results}
\label{sec:results}

\paragraph{Reading the results.}
All percentages are shares of individual replies, turns, or member endpoints, not of questions or final
committee answers. Differences between percentages are in percentage points (pp). Replies within a question,
path, or tone assignment are not independent, so we report descriptive rates; the per-cell denominators are
in the tables. We abbreviate the agreement levels as SA (strongly agree), LA (leaning agree), LD (leaning
disagree), and SD (strongly disagree). Size in result tables is the number of members who debate; two-member committees pair two of the three
members.

\subsection{Layer A: tone shifts reported agreement, moderated by model choice}
\label{sec:grid}
Friendly instructions produce the most strong-agreement labels and hostile instructions the fewest, with
neutral instructions close to friendly ones (Table~\ref{tab:rd-agreement}). Across the six settings, the
share of replies labeled strongly agree is \rd{A/friendly/sa/min}--\rd{A/friendly/sa/max}\% under friendly
instructions and \rd{A/hostile/sa/min}--\rd{A/hostile/sa/max}\% under hostile ones. Without any tone
instruction, members already report strong agreement with most incoming contributions
(\rd{A/neutral/sa/min}--\rd{A/neutral/sa/max}\%), so the default debate is largely consensual.

Model choice changes both the size of the tone effect and what replaces strong agreement.
The friendly--hostile difference in strong agreement is largest when all members share one model
(\rd{A/dyad/same/gap}pp with two members and \rd{A/triad/same/gap}pp with three) and smallest in mixed-family
committees (\rd{A/dyad/mixed/gap}pp and \rd{A/triad/mixed/gap}pp). Under hostile instructions, same-model
members mostly move to leaning agreement (\rd{A/dyad/same/hostile/la}\% and \rd{A/triad/same/hostile/la}\%),
whereas mixed-family members report leaning disagreement far more often (\rd{A/dyad/mixed/hostile/ld}\% and
\rd{A/triad/mixed/hostile/ld}\%, versus \rd{A/dyad/same/hostile/ld}\% and \rd{A/triad/same/hostile/ld}\% for
the same model). Strong disagreement is almost never reported: \rd{A/sd/total} of \rd{A/replies/total}
replies in total. Hostility therefore mainly qualifies agreement among similar models, and turns it into
qualified disagreement more often among different models. Committee size has little effect: two- and
three-member committees in the same configuration show similar distributions.

\begin{table}[t]
\centering\small
\begin{tabular}{lllrrrrrrrrrrrr}
\toprule
 & & & \multicolumn{4}{c}{Friendly} & \multicolumn{4}{c}{Neutral} & \multicolumn{4}{c}{Hostile}\\
\cmidrule(lr){4-7}\cmidrule(lr){8-11}\cmidrule(lr){12-15}
Size & Setup & Models & $n$ & SA & LA & LD & $n$ & SA & LA & LD & $n$ & SA & LA & LD\\
\midrule
Two & Same model & Terra $\times$3 & 300 & 92.3 & 2.0 & 5.7 & 300 & 87.3 & 0.3 & 12.3 & 300 & 16.7 & 71.3 & 12.0\\
 & Same family & Luna, Terra, Sol & 300 & 81.0 & 6.0 & 13.0 & 300 & 69.7 & 6.7 & 23.7 & 300 & 19.3 & 57.7 & 23.0\\
 & Mixed family & Terra, Qwen, Inkling & 300 & 70.7 & 15.3 & 14.0 & 300 & 66.7 & 12.0 & 21.0 & 300 & 9.7 & 46.0 & 43.7\\
\midrule
Three & Same model & Terra $\times$3 & 574 & 96.3 & 0.3 & 3.3 & 566 & 92.4 & 2.8 & 4.8 & 559 & 14.0 & 77.6 & 8.4\\
 & Same family & Luna, Terra, Sol & 574 & 83.6 & 3.7 & 12.7 & 566 & 76.3 & 4.2 & 19.4 & 559 & 15.2 & 66.7 & 18.1\\
 & Mixed family & Terra, Qwen, Inkling & 574 & 70.7 & 14.6 & 14.6 & 566 & 64.7 & 14.1 & 21.0 & 559 & 9.5 & 51.2 & 38.5\\
\bottomrule
\end{tabular}

\caption{\textbf{Layer A: self-reported agreement by turn tone (\% of replies).} $n$ is the number of unique
replies at turns with that tone. SD is omitted because it is nearly empty (\rd{A/sd/total} of
\rd{A/replies/total} replies). Same model: three GPT-5.6 Terra members; same family: GPT-5.6 Luna, Terra, and Sol; mixed family: GPT-5.6 Terra, Qwen3.8-27B, and Inkling.}
\label{tab:rd-agreement}
\end{table}

\subsection{Layer B: the reply text largely corroborates the self-reports}
\label{sec:text-results}
The external judge's ratings agree with the members' self-labels in \rd{B/exact/min}--\rd{B/exact/max}\% of
sampled replies, with Cohen's $\kappa$ between \rd{B/kappa/min} and \rd{B/kappa/max}
(Table~\ref{tab:rd-text}). Strong-agreement labels are almost always confirmed by the text
(\rd{B/triad/mixed/satosa}--\rd{B/dyad/same/satosa}\%). The labels in between are less consistent: the judge
reads \rd{B/latosa/min}--\rd{B/latosa/max}\% of replies labeled leaning agree as strongly agree, and
\rd{B/triad/same/ldtosd}--\rd{B/dyad/mixed/ldtosd}\% of replies labeled leaning disagree as strongly
disagree.

Consistency also depends on the model choice. It is lowest in the mixed-family committees
($\kappa={}$\rd{B/dyad/mixed/kappa} and \rd{B/triad/mixed/kappa}), where the judge assigns a more disagreeing
label than the member in \rd{B/dyad/mixed/judgemoredisagree}\% (two members) and
\rd{B/triad/mixed/judgemoredisagree}\% (three members) of replies. In same-model committees, mismatches
point mostly the other way: the judge reads more agreement than the member reports in
\rd{B/dyad/same/judgemoreagree}\% and \rd{B/triad/same/judgemoreagree}\% of replies, and less in only
\rd{B/dyad/same/judgemoredisagree}\% and \rd{B/triad/same/judgemoredisagree}\%.
Thus the tone pattern in \layerA{} is not
merely a labeling artifact: it appears in what members write.
This is agreement with a peer's message, not a change in the member's answer to the original question,
which we turn to next. On the 50-reply grounding sample, a second judge (Claude Opus 5.5) agrees with these ratings
on \rd{G/opus/same} of \rd{G/opus/n} replies (Section~\ref{sec:grounding}; Appendix~\ref{app:judge}).

\begin{table}[t]
\centering\small
\begin{tabular}{lllrrrrrrrr}
\toprule
 & & & & & & \multicolumn{5}{c}{Judge label given self-label (\%)}\\
\cmidrule(lr){7-11}
Size & Setup & Models & $n$ & Exact & $\kappa$ & SA$\to$SA & LA$\to$SA & LA$\to$LA & LD$\to$LD & LD$\to$SD\\
\midrule
Two & Same model & Terra $\times$3 & 667 & 91.3 & 0.81 & 99.8 & 27.7 & 72.3 & 80.3 & 13.6\\
 & Same family & Luna, Terra, Sol & 667 & 89.4 & 0.81 & 99.0 & 26.5 & 69.5 & 83.2 & 12.0\\
 & Mixed family & Terra, Qwen, Inkling & 667 & 82.2 & 0.72 & 99.4 & 24.3 & 59.2 & 71.4 & 21.4\\
\midrule
Three & Same model & Terra $\times$3 & 731 & 92.2 & 0.83 & 99.6 & 24.7 & 75.3 & 86.0 & 7.0\\
 & Same family & Luna, Terra, Sol & 731 & 91.4 & 0.85 & 98.1 & 19.1 & 79.8 & 85.8 & 11.3\\
 & Mixed family & Terra, Qwen, Inkling & 731 & 81.8 & 0.71 & 97.5 & 25.8 & 59.8 & 72.8 & 19.5\\
\bottomrule
\end{tabular}

\caption{\textbf{Layer B: judge labels versus self-labels.} $n$ is the number of judged replies with both
labels; Exact is the share with identical labels; $\kappa$ is Cohen's kappa. The right block gives the
judge's label distribution within each self-label, e.g., LA$\to$SA is the share of replies self-labeled
leaning agree that the judge rates strongly agree. Self-label SD is omitted (\rd{B/triad/mixed/n} sampled
three-member mixed-family replies include two; no other setting has any). Same model: three GPT-5.6 Terra members; same family: GPT-5.6 Luna, Terra, and Sol; mixed family: GPT-5.6 Terra, Qwen3.8-27B, and Inkling.}
\label{tab:rd-text}
\end{table}

\subsection{Layer C: positions rarely change, and mostly when a member concedes}
\label{sec:persistence}
Despite the large differences in reported agreement, members seldom change their own answers. Across the
sampled two-member turns, members switch their option in \rd{C/same/option/all}\% (same model),
\rd{C/family/option/all}\% (same family), and \rd{C/mixed/option/all}\% (mixed family) of comparisons with
their previous participation (Table~\ref{tab:rd-position}).

The breakdown by both agreement labels shows where the switches occur. When the peer's preceding reply is
labeled leaning disagree, i.e., the peer pushed back against the member, and the member then reports strong
agreement with that pushback, it switches options in \rd{C/same/ld/sa/option}\% (same model), \rd{C/family/ld/sa/option}\% (same family),
and \rd{C/mixed/ld/sa/option}\% (mixed family) of cases, and the judge finds a changed conclusion at similar rates. When the
member instead keeps disagreeing (LD after LD), it switches in at most \rd{C/ld/ld/option/max}\% of cases.
After a peer's strong agreement, options almost never change. Qualified concessions sit in between and vary by
model: after a peer's leaning disagreement, members that report leaning agreement switch in
\rd{C/family/ld/la/option}\% of same-family and \rd{C/mixed/ld/la/option}\% of mixed-family cases.
These labels are observed rather than assigned, so the rows describe which interactions accompany a change,
not what causes it. Appendix~\ref{app:examples} shows one such concession, where the member accepts the peer's
argument and switches to another option.

\begin{table}[t]
\centering\small
\begin{tabular}{llrrrrrrrrr}
\toprule
 & & \multicolumn{3}{c}{Same model} & \multicolumn{3}{c}{Same family} & \multicolumn{3}{c}{Mixed family}\\
 & & \multicolumn{3}{c}{(Terra $\times$3)} & \multicolumn{3}{c}{(Luna, Terra, Sol)} & \multicolumn{3}{c}{(Terra, Qwen, Inkling)}\\
\cmidrule(lr){3-5}\cmidrule(lr){6-8}\cmidrule(lr){9-11}
Peer & Own & $n$ & Opt. & Concl. & $n$ & Opt. & Concl. & $n$ & Opt. & Concl.\\
\midrule
SA & SA & 323 & 0.6 & 0.0 & 275 & 0.0 & 0.0 & 244 & 0.4 & 0.0\\
SA & LA & 113 & 1.8 & 0.0 & 93 & 0.0 & 0.0 & 79 & 1.3 & 0.0\\
SA & LD & 2 & 50.0 & 50.0 & 4 & 75.0 & 75.0 & 14 & 7.1 & 7.1\\
LA & SA & 103 & 3.9 & 1.0 & 97 & 1.1 & 1.0 & 66 & 4.5 & 3.0\\
LA & LA & 45 & 0.0 & 0.0 & 44 & 2.3 & 0.0 & 48 & 6.2 & 0.0\\
LA & LD & 4 & 33.3 & 25.0 & 18 & 0.0 & 0.0 & 37 & 2.8 & 2.7\\
LD & SA & 16 & 100.0 & 93.8 & 19 & 100.0 & 89.5 & 20 & 55.0 & 55.0\\
LD & LA & 1 & -- & 100.0 & 14 & 53.8 & 50.0 & 40 & 17.5 & 15.0\\
LD & LD & 60 & 1.9 & 1.7 & 103 & 3.1 & 0.0 & 116 & 1.8 & 1.7\\
\bottomrule
\end{tabular}

\caption{\textbf{Layer C: turn-level position change in two-member committees (\%).} Rows give the agreement
label of the peer's preceding reply and the member's own label in the current turn; combinations with fewer
than five turns in every configuration are omitted. $n$ is the number of sampled turns; Opt.\ is the share with
a changed option; Concl.\ is the share whose position text the judge rates as conclusion changed. Small cells (e.g., $n<20$) are unstable. Each two-member committee pairs
two of the three listed models.}
\label{tab:rd-position}
\end{table}

\subsection{Layer D: the held option stays near certain, while endorsement of the earlier reasoning softens}
\label{sec:logprob}
\paragraph{Option probabilities (D1).}
For the open-weight members of the mixed-family two-member committees (Qwen3.8-27B and Inkling), the probability
of the option a member already holds is close to saturation and barely moves within a turn
(Table~\ref{tab:rd-d1}). Across all later turns, Qwen3.8-27B's probability of its held option goes from
\rd{D1/qwen/later/before}\% to \rd{D1/qwen/later/after}\% (\rd{D1/qwen/later/change}pp), and Inkling's from
\rd{D1/inkling/later/before}\% to \rd{D1/inkling/later/after}\% (\rd{D1/inkling/later/change}pp). Pushback does not
change this picture: after a peer's leaning or strong disagreement, the change is \rd{D1/qwen/later_pushback/change}pp
for Qwen3.8-27B (\rd{D1/qwen/later_pushback/up} reads up and \rd{D1/qwen/later_pushback/down} down by more than
1pp, of \rd{D1/qwen/later_pushback/n}) and \rd{D1/inkling/later_pushback/change}pp for Inkling. When the member
keeps disagreeing (LD after LD), the held option loses little support (\rd{D/qwen/ld/ld/d1_pp}pp and
\rd{D/inkling/ld/ld/d1_pp}pp). The few cells with larger increases, where the member reports agreeing with a peer's
pushback, contain fewer than 15 comparisons each.
We exclude each member's first turn: there, the earlier read saw only the question and none of the member's own
answer, and the held option's probability jumps from \rd{D1/qwen/first/before}\% to \rd{D1/qwen/first/after}\%
for Qwen3.8-27B (\rd{D1/inkling/first/before}\% to \rd{D1/inkling/first/after}\% for Inkling), which reflects
having stated an answer at all rather than the debate.

\begin{table}[t]
\centering\small
\begin{tabular}{llrrrrrrrr}
\toprule
 & & \multicolumn{4}{c}{Qwen3.8-27B} & \multicolumn{4}{c}{Inkling}\\
\cmidrule(lr){3-6}\cmidrule(lr){7-10}
Peer & Own & $n$ & Before (\%) & After (\%) & $\Delta p$ (pp) & $n$ & Before (\%) & After (\%) & $\Delta p$ (pp)\\
\midrule
SA & SA & 52 & 97.7 & 99.0 & +1.3 & 42 & 100.0 & 100.0 & 0.0\\
SA & LA & 15 & 88.4 & 93.7 & +5.4 & 14 & 100.0 & 100.0 & 0.0\\
SA & LD & 0 & -- & -- & -- & 6 & 99.8 & 100.0 & +0.2\\
LA & SA & 11 & 92.0 & 97.2 & +5.2 & 16 & 100.0 & 99.4 & $-$0.6\\
LA & LA & 11 & 84.1 & 93.9 & +9.7 & 8 & 96.8 & 99.9 & +3.1\\
LA & LD & 2 & 91.8 & 83.2 & $-$8.6 & 8 & 93.0 & 99.7 & +6.7\\
LD & SA & 6 & 75.5 & 90.9 & +15.4 & 5 & 74.7 & 92.1 & +17.4\\
LD & LA & 14 & 85.7 & 93.7 & +8.0 & 7 & 80.6 & 96.9 & +16.3\\
LD & LD & 22 & 91.2 & 90.9 & $-$0.3 & 29 & 93.9 & 87.9 & $-$6.0\\
\bottomrule
\end{tabular}

\caption{\textbf{Layer D1: probability of the held option before consecutive turns}, for the open-weight members
(Qwen3.8-27B and Inkling) of the mixed-family two-member committees; their peer is the other open-weight model or
GPT-5.6 Terra. Later turns only: both reads include the member's own earlier answer. $n$ is the number of
comparisons. Before and After are the probabilities of the option the member held at its previous turn, read
before that turn and before the current one; $\Delta p$ is their difference (pp). Values average within
questions, then weight questions equally. Rows are those of Table~\ref{tab:rd-d2}: combinations with at least five D2 pairs for either model.}
\label{tab:rd-d1}
\end{table}

\paragraph{Endorsement of the earlier position (D2).}
The endorsement probe, which holds everything but the incoming message fixed, does respond to the interaction
(Table~\ref{tab:rd-d2}). After a peer's strong agreement, the argument and the filler give nearly the same
endorsement of the member's earlier position (\rd{D/qwen/sa/sa/d2_filler} versus \rd{D/qwen/sa/sa/d2_argument} for
Qwen3.8-27B on the 1--7 scale). After pushback, Qwen3.8-27B's endorsement drops by about one step: when both labels
are leaning disagree, its expected rating falls from \rd{D/qwen/ld/ld/d2_filler} to \rd{D/qwen/ld/ld/d2_argument}.
In concrete terms, its most likely rating moves from G (\emph{completely agree}) to F (\emph{mostly agree}) in
\rd{D2/qwen/pushback/GtoF} of \rd{D2/qwen/pushback/n} pairs after pushback, compared with
\rd{D2/qwen/peer_sa/GtoF} of \rd{D2/qwen/peer_sa/n} after a peer's strong agreement. Qwen3.8-27B thus still agrees
with its earlier position, only less than completely. Inkling's endorsement barely changes (its most likely rating
stays at G in \rd{D2/inkling/pushback/GtoG} of \rd{D2/inkling/pushback/n} pairs after pushback).

D1 and D2 measure different things and point the same way. The member's conclusion, its chosen option, stays
almost certain, while its endorsement of the exact earlier wording, reasons included, softens by a step after
pushback. These are probabilities of a model's next answer in a specific
context, not measurements of lasting beliefs.

\begin{table}[t]
\centering\small
\begin{tabular}{llrrrrrrrr}
\toprule
 & & \multicolumn{4}{c}{Qwen3.8-27B} & \multicolumn{4}{c}{Inkling}\\
\cmidrule(lr){3-6}\cmidrule(lr){7-10}
Peer & Own & $n$ & Filler & Argument & $\Delta$ & $n$ & Filler & Argument & $\Delta$\\
\midrule
SA & SA & 69 & 6.84 & 6.86 & +0.02 & 62 & 6.99 & 6.98 & $-$0.01\\
SA & LA & 18 & 6.83 & 6.86 & +0.03 & 18 & 7.00 & 7.00 & 0.00\\
SA & LD & 0 & -- & -- & -- & 8 & 7.00 & 7.00 & 0.00\\
LA & SA & 12 & 6.82 & 6.56 & $-$0.27 & 21 & 6.97 & 6.84 & $-$0.13\\
LA & LA & 18 & 6.75 & 5.94 & $-$0.80 & 8 & 7.00 & 6.84 & $-$0.16\\
LA & LD & 6 & 6.65 & 6.13 & $-$0.52 & 11 & 6.92 & 6.89 & $-$0.04\\
LD & SA & 6 & 6.75 & 5.72 & $-$1.03 & 5 & 6.99 & 6.80 & $-$0.19\\
LD & LA & 19 & 6.76 & 5.58 & $-$1.18 & 11 & 7.00 & 6.87 & $-$0.13\\
LD & LD & 29 & 6.74 & 5.81 & $-$0.92 & 36 & 6.97 & 6.84 & $-$0.13\\
\bottomrule
\end{tabular}

\caption{\textbf{Layer D2: endorsement of the earlier position, filler versus peer argument}, for the same
open-weight members as Table~\ref{tab:rd-d1}. Expected rating of the member's earlier position on a 1--7 scale
(1 = completely disagree, 7 = completely agree) when the incoming message is neutral filler or the peer's actual
argument; $\Delta$ is argument minus filler (negative: lower endorsement after the argument). $n$ is the number of
argument--filler pairs; values average within questions, then weight questions equally.}
\label{tab:rd-d2}
\end{table}

\subsection{Layer E: final positions rarely differ from initial ones}
\label{sec:endpoint-results}
Over a complete debate path, the selected option is unchanged in \rd{E/unchanged/min}--\rd{E/unchanged/max}\%
of member endpoints across the six settings (Table~\ref{tab:rd-endpoint}); no single member changes its option
in more than \rd{E/member/option/max}\% of its endpoints. The judge finds a changed conclusion in at most
\rd{E/conclusion/max}\% of sampled endpoint pairs in any setting.
Three-member debates, which are longer and expose each member to two peers, do not produce more change than
two-member debates. Over a complete debate, then, members largely retain their initial answers: expressed
disagreement, including the hostile turns' qualified disagreement, rarely turns into a revised position.

\begin{table}[t]
\centering\small
\begin{tabular}{lllrrrr}
\toprule
 & & & \multicolumn{2}{c}{Option} & \multicolumn{2}{c}{Position text (judge)}\\
\cmidrule(lr){4-5}\cmidrule(lr){6-7}
Size & Setup & Member & $n$ & Changed & $n$ & Concl.\\
\midrule
Two & Same model & GPT-5.6 Terra (A) & 192 & 6.2 & 178 & 5.1\\
 &  & GPT-5.6 Terra (B) & 196 & 5.6 & 176 & 4.5\\
 &  & GPT-5.6 Terra (C) & 200 & 4.5 & 180 & 4.4\\
\addlinespace
 & Same family & GPT-5.6 Luna & 192 & 5.7 & 178 & 3.4\\
 &  & GPT-5.6 Terra & 194 & 12.4 & 176 & 10.8\\
 &  & GPT-5.6 Sol & 200 & 5.5 & 180 & 5.0\\
\addlinespace
 & Mixed family & GPT-5.6 Terra & 196 & 6.6 & 178 & 5.1\\
 &  & Qwen3.8-27B & 200 & 7.0 & 176 & 4.0\\
 &  & Inkling & 200 & 3.0 & 180 & 3.3\\
\midrule
Three & Same model & GPT-5.6 Terra (A) & 564 & 5.0 & 194 & 5.2\\
 &  & GPT-5.6 Terra (B) & 558 & 2.5 & 184 & 2.2\\
 &  & GPT-5.6 Terra (C) & 554 & 5.1 & 184 & 6.0\\
\addlinespace
 & Same family & GPT-5.6 Luna & 558 & 3.6 & 194 & 2.1\\
 &  & GPT-5.6 Terra & 553 & 6.0 & 184 & 6.5\\
 &  & GPT-5.6 Sol & 542 & 1.5 & 184 & 1.6\\
\addlinespace
 & Mixed family & GPT-5.6 Terra & 563 & 3.0 & 194 & 2.6\\
 &  & Qwen3.8-27B & 566 & 4.8 & 184 & 3.8\\
 &  & Inkling & 542 & 2.2 & 184 & 2.2\\
\bottomrule
\end{tabular}

\caption{\textbf{Layer E: change from each member's initial position to its last participation (\%).}
Option changes use every available path endpoint; Concl.\ is the share of endpoint pairs, sampled before
generation, whose position text the judge rates as conclusion changed, and each block has its own $n$. Same-model members are three distinct GPT-5.6 Terra instances (A--C).}
\label{tab:rd-endpoint}
\end{table}

\subsection{Supplementary study: forced disagreement rarely changes positions}
\label{sec:forced-results}
Instructing the challenger to oppose the incoming message works as intended at the level of expressed
disagreement (Table~\ref{tab:rd-forced}). Every forced challenger reply is labeled leaning or strongly disagree,
\rd{F/all/forced/challenger/sd}\% strongly, whereas in the natural debates \rd{F/all/natural/challenger/sa}\% of
the same replies report strong agreement. The judge reads the forced text as more disagreeing still: it rates
\rd{F/all/forced/challenger/ldtosd}\% of the challengers' leaning-disagree replies as strongly disagree. The
recipients respond in kind: \rd{F/all/forced/recipient/ld}\% of their replies report leaning disagreement, compared
with \rd{F/all/natural/recipient/ld}\% in the natural debates, and strong agreement falls from
\rd{F/all/natural/recipient/sa}\% to \rd{F/all/forced/recipient/sa}\%.

Positions, however, barely respond, which is consistent with our findings in Layer C. Recipients change their option
in \rd{F/all/forced/option}\% of forced cases, compared with \rd{F/all/natural/option}\% in the natural debates, and
the judge finds a changed conclusion in \rd{F/all/forced/conclusion}\% versus \rd{F/all/natural/conclusion}\%. The
probability reads show the same split as in the main study (Table~\ref{tab:rd-forced-d}). The forced argument lowers
Qwen3.8-27B's endorsement of its earlier position from \rd{F/all/forced/qwen/d2_filler} to
\rd{F/all/forced/qwen/d2_argument}, more than twice the drop in the natural debate (\rd{F/all/natural/qwen/d2_filler}
to \rd{F/all/natural/qwen/d2_argument}), while Inkling's endorsement barely moves. Its probability of the held
option changes by \rd{F/all/forced/qwen/d1_pp}pp, compared with \rd{F/all/natural/qwen/d1_pp}pp in the natural
debate (Inkling: \rd{F/all/forced/inkling/d1_pp}pp versus \rd{F/all/natural/inkling/d1_pp}pp). Sustained,
substantive opposition thus changes what the recipient says and, for Qwen3.8-27B, how strongly it endorses its
earlier wording, but leaves its answer largely in place.

\begin{table}[t]
\centering\small
\setlength{\tabcolsep}{4.5pt}
\begin{tabular}{lrrrrrrrrrrr}
\toprule
 & & \multicolumn{4}{c}{Challenger's label} & \multicolumn{4}{c}{Recipient's label} & \multicolumn{2}{c}{Recipient's position}\\
\cmidrule(lr){3-6}\cmidrule(lr){7-10}\cmidrule(lr){11-12}
Arm & $n$ & SA & LA & LD & SD & SA & LA & LD & SD & Opt. & Concl.\\
\midrule
Natural & 300 & 52.3 & 20.7 & 26.0 & 1.0 & 50.0 & 24.3 & 25.7 & 0.0 & 6.4 & 5.0\\
Forced & 300 & 0.0 & 0.0 & 73.3 & 26.7 & 6.3 & 13.3 & 76.0 & 4.3 & 7.7 & 6.7\\
\bottomrule
\end{tabular}

\caption{\textbf{Forced disagreement in mixed-family two-member committees (GPT-5.6 Terra, Qwen3.8-27B,
Inkling) (\%).} \emph{Forced}: the challenger is privately instructed to oppose the message it answers;
\emph{Natural}: the main study's continuation of the same path, stopped at the same turn. $n$ is the number of
cases, one per question, pair, and intervention point. The label columns give the self-reported agreement
distribution of the challenger's reply and of the recipient's reply. Opt.\ and Concl.\ are the recipient's
option change and the judge's conclusion-changed rating, from its position before the challenger's reply to its
own reply.}
\label{tab:rd-forced}
\end{table}

\begin{table}[t]
\centering\small
\begin{tabular}{lrrrr}
\toprule
 & \multicolumn{2}{c}{Qwen3.8-27B} & \multicolumn{2}{c}{Inkling}\\
\cmidrule(lr){2-3}\cmidrule(lr){4-5}
Arm & D1 $\Delta p$ (pp) & D2: filler $\to$ argument & D1 $\Delta p$ (pp) & D2: filler $\to$ argument\\
\midrule
Natural & +4.8 & 6.76 $\to$ 6.26 ($-$0.50) & +0.9 & 6.99 $\to$ 6.89 ($-$0.10)\\
Forced & $-$2.5 & 6.76 $\to$ 5.54 ($-$1.23) & $-$3.9 & 6.99 $\to$ 6.79 ($-$0.20)\\
\bottomrule
\end{tabular}

\caption{\textbf{Forced disagreement: probability reads for open-weight recipients.} D1 $\Delta p$ is the change in
probability of the held option (pp), as in Table~\ref{tab:rd-d1}; it uses the \rd{F/all/qwen/d1n} cases per model in
which the recipient had already replied once before the intervention, since otherwise its earlier read saw only
the question. D2 gives the expected endorsement rating of the recipient's earlier position after filler and after
the argument (1--7 scale), with their difference in parentheses, as in Table~\ref{tab:rd-d2}, over
\rd{F/all/qwen/d2n} cases per model. Values average within questions first.}
\label{tab:rd-forced-d}
\end{table}

\section{Implications and Limitations}
\label{sec:limitations}
\paragraph{Implications for agent design.}
Debating the same question does not, by itself, produce human-like brainstorming. Tone instructions change how
much agreement members report, but members switch options in only \rd{C/same/option/all}--\rd{C/family/option/all}\%
of turns, and over a complete debate they keep their initial option in \rd{E/unchanged/min}--\rd{E/unchanged/max}\% of
comparisons. Members do change their position when a peer's argument persuades them: after a peer's pushback, members
that then report strong agreement switch in \rd{C/mixed/ld/sa/option}--\rd{C/same/ld/sa/option}\% of cases. In the
concession example (Appendix~\ref{app:examples}), the member switches because the peer points out how the survey
question is ordinarily read, a consideration missing from its own reasoning. Such missing context is rare when every
member answers from the same prompt and the same knowledge, and likely more common when agents act in different
environments or use different tools. Forcing disagreement does not substitute for it: recipients of forced opposition
report leaning disagreement in \rd{F/all/forced/recipient/ld}\% of cases, yet change their option in only
\rd{F/all/forced/option}\%, compared with \rd{F/all/natural/option}\% without the intervention. When all agents share
the same context, debate rarely changes what they hold, even when they are made to disagree, and mixing model
families changes what members say but not what they hold: option changes stay at
\rd{C/same/option/all}--\rd{C/family/option/all}\% of turns in every configuration. Multi-agent debate systems should
therefore aim to give members different context, such as different information, environments, or tools, rather than
rely on differences in model behavior; we leave a direct test of this to future work.

\paragraph{Limitations: setups and data.}
We evaluate a specific set of committee configurations (same-model, same-family, and mixed-family committees of
two or three members, built from five models) on \rd{questions} questions from GlobalOpinionQA. The results may
differ for other agent setups, models, debate protocols, and datasets. Each tone uses a single instruction wording,
and all questions are in English.

\paragraph{Limitations: LLM judges.}
We use an LLM judge to measure agreement in reply text for \layerB{} and changes in position text for \layerC{}
(and \layerE{}). Comparing two short texts is a simple task for current LLMs, but the judge can still make errors,
which would carry into these measurements. A second LLM judge agrees with it on \rd{G/opus/same} of \rd{G/opus/n}
sampled replies for \layerB{} and on whether the conclusion changed in \rd{G/ce/same} of \rd{G/ce/n} sampled
comparisons for \layerC{} and \layerE{} (Section~\ref{sec:grounding}).

\section{Conclusion}We asked whether the disagreement that LLM committee members express during debate reflects changes in their
own positions, and answered it with five measurements of the same debates.
Tone instructions strongly change the agreement members report (\layerA{}), and the reply text largely
corroborates those reports (\layerB{}), with model choice shaping both: similar models mostly qualify their
agreement, while models from different families more often express qualified disagreement.
Yet members rarely change their answers at a debate turn (\layerC{}); when they do, it is mostly when they
explicitly concede to a peer's pushback. For open-weight members, a peer's argument can lower the endorsement of
their earlier wording, while the probability of the option they hold stays near certain (\layerD{}).
Over a complete debate, members keep their initial option in more than nine of ten comparisons (\layerE{}).

Expressed disagreement and position revision are therefore different outcomes, and measuring one does not
measure the other. Evaluations of multi-agent debate that rely on what agents say about agreement, or on how
much they argue, should also track what the agents end up holding.

\bibliographystyle{plainnat}
\bibliography{shared/refs}

\appendix
\raggedbottom

\section{Question bank}
\label{app:questions}
We screened GlobalOpinionQA~\citep{durmus2023globalopinions} records in the dataset's fixed order.
Three LLM classifiers (GPT-5.5, Claude Opus 4.8, and Gemini 3.5 Flash) each labeled whether a candidate
asks for an opinion or for a report of the respondent's personal experiences, habits, or identity, and we
kept candidates that at least two classifiers labeled as opinion questions. We then removed questions whose
answer depends on a national context the question does not state, such as an unspecified ``your country,''
``the government,'' or ``our daily lives''; questions that name a country, or ask about institutions in
general, were kept. Question wording, options, and option order are unchanged from the dataset, including
non-substantive options such as ``Don't know'' or ``Refused.'' This process produced a bank of 60 questions.
The study uses the \rd{questions} questions below, drawn uniformly at random from the bank before any debate;
all configurations share them.

\begin{enumerate}\small
\item Please tell us if you strongly agree, agree, disagree, or strongly disagree with the following statements: It is a man’s job to earn money and a women’s job to take care of home and family\\ \textit{Options:} A: Strongly agree; B: Agree; C: Disagree; D: Strongly disagree; E: Don't know; F: No answer/refused
\item Would you say that each of the following has done a very good, somewhat good, somewhat bad, or very bad job dealing with the coronavirus outbreak? d. The European Union\\ \textit{Options:} A: Very good; B: Somewhat good; C: Somewhat bad; D: Very bad; E: DK/Refused
\item Which comes closer to describing your view?...It's good that American ideas and customs are spreading around the world OR It's bad that American ideas and customs are spreading around the world\\ \textit{Options:} A: It's good that American ideas and customs are spreading around the world; B: It's bad that American ideas and customs are spreading around the world
\item And of the items you mentioned as very important, which is the most important for the U.S to do...build a strong relationship with China, be tough with China on economic and trade issues, promote human rights in China, advocate for more freedom for Tibet, or continue selling arms to Taiwan?\\ \textit{Options:} A: Build a strong relationship with China; B: Be tough with China on economic/trade issues; C: Promote human rights in China; D: Advocate for more freedom for Tibet; E: Continue selling arms to Taiwan; F: None
\item I am going to name a number of organizations. For each one, could you tell me how much confidence you have in them: is it a great deal of confidence, quite a lot of confidence, not very much confidence or none at all? Universities\\ \textit{Options:} A: A great deal; B: Quite a lot; C: Not very much; D: None at all; E: Don't know; F: No answer; G: Missing; Not available
\item Which comes closer to describing your view?...I like American ways of doing business, OR I dislike American ways of doing business\\ \textit{Options:} A: I like American ways of doing business; B: I dislike American ways of doing business
\item What is most responsible for Muslim nations' lack of prosperity? Is it the policies of the U.S. and other western nations, the lack of democracy in the Muslim world, the lack of education in the Muslim world, Islamic fundamentalism, or corruption in the government of Muslim nations?\\ \textit{Options:} A: The policies of the U.S. and other western nations; B: Lack of democracy; C: Lack of education; D: Islamic fundamentalism; E: Corruption in the governments of Muslim nations; F: Other (VOL)
\item Generally speaking, how much do you think the United States can trust Japan...a great deal, a fair amount, not too much, or not at all?\\ \textit{Options:} A: Great deal; B: Fair amount; C: Not too much; D: Not at all
\item Thinking about conditions in Egypt in recent years, what has concerned you most?...Lack of economic prosperity, Lack of democracy, Corruption, the potential for political instability, or the banning of religious parties?\\ \textit{Options:} A: Lack of economic prosperity; B: Lack of democracy; C: Corruption; D: The potential for political instability; E: The banning of religious parties; F: Other (VOL)
\item As I read some proposed policies of President Donald Trump, please tell me if you approve or disapprove of each one. b. build a wall on the border between the U.S. and Mexico\\ \textit{Options:} A: Approve; B: Disapprove; C: DK/Refused
\item I am going to name a number of organizations. For each one, could you tell me how much confidence you have in them: is it a great deal of confidence, quite a lot of confidence, not very much confidence or none at all? The Churches (mosque, temple etc.)\\ \textit{Options:} A: A great deal; B: Quite a lot; C: Not very much; D: None at all; E: Don't know; F: No answer; G: Other missing; Multiple answers Mail (EVS)
\item Please tell me if you have a very favorable, somewhat favorable, somewhat unfavorable, or very unfavorable opinion of...Muslims\\ \textit{Options:} A: Very favorable; B: Somewhat favorable; C: Somewhat unfavorable; D: Very unfavorable; E: Never heard of
\item Here is a list of four dangers in the world today. In your opinion, which one of these poses the greatest threat to the world?\\ \textit{Options:} A: Spread of nuclear weapons; B: AIDS and other infectious diseases; C: Pollution and other environmental problems; D: Growing gap between the rich and poor
\item Do you agree or disagree with the following statement? A university education is more important for a boy than for a girl. Do you completely agree, mostly agree, mostly disagree or completely disagree?\\ \textit{Options:} A: Completely agree; B: Mostly agree; C: Mostly disagree; D: Completely disagree
\item I am going to name a number of organizations. For each one, could you tell me how much confidence you have in them: is it a great deal of confidence, quite a lot of confidence, not very much confidence or none at all? Women's organizations\\ \textit{Options:} A: A great deal; B: Quite a lot; C: Not very much; D: None at all; E: Don't know; F: No answer; G: Missing; Not available
\item Do you think \_\_\_\_ plays a more important role in the world today compared to 10 years ago, a less important role, or about as important a role in the world as it did 10 years ago? f. France\\ \textit{Options:} A: More important role; B: Less important role; C: As important as 10 years ago; D: DK/Refused
\item I am going to name a number of organizations. For each one, could you tell me how much confidence you have in them: is it a great deal of confidence, quite a lot of confidence, not very much confidence or none at all? Elections\\ \textit{Options:} A: A great deal; B: Quite a lot; C: Not very much; D: None at all; E: Don't know; F: No answer; G: Missing; Unknown
\item How much, if at all, do you think the U.S. benefits from being a member of each of the following organizations? b. The United Nations\\ \textit{Options:} A: A great deal; B: A fair amount; C: Not too much; D: Not at all; E: DK/Refused
\item Do you think people should be able to say these types of things publically OR the government should be able to prevent people from saying these things in some circumstances. b. statements that are offensive to minority groups\\ \textit{Options:} A: People should be able to say these things publically; B: Government should be able to prevent people from saying these things; C: DK/Refused
\item Now I'm going to read a list of political leaders. For each, tell me how much confidence you have in each leader to do the right thing regarding world affairs - a lot of confidence, some confidence, not too much confidence, or no confidence at all?...U.S. Presidential candidate Hillary Clinton\\ \textit{Options:} A: A lot of confidence; B: Some confidence; C: Not too much confidence; D: No confidence at all; E: DK/Refused
\item Do you think US policies in the Middle East make the region more stable or less stable?\\ \textit{Options:} A: More stable; B: Less stable; C: No difference (VOL)
\item Thinking about how people around the world view Mexico these days, do you think Mexico is well regarded or poorly regarded?\\ \textit{Options:} A: Well regarded; B: Poorly regarded; C: Neither/both (VOL)
\item Do you support or oppose the U.S. military actions against the Islamic militant group in Iraq and Syria known as ISIS?\\ \textit{Options:} A: Support; B: Oppose
\item Do you personally believe that homosexuality is morally acceptable, morally unacceptable, or is it not a moral issue?\\ \textit{Options:} A: Morally acceptable; B: Morally unacceptable; C: Not a moral issue; D: Depends on the situation (VOL)
\item In general, do you think the Israeli government favors or opposes democracy in the Middle East?\\ \textit{Options:} A: Favors; B: Opposes; C: Both/neither (VOL)
\item Which comes closer to describing your view?...I admire the United States for its technological and scientific advances, OR I do not admire the United States for its technological and scientific advances\\ \textit{Options:} A: I admire the US for its technological/ scientific advances; B: I do not admire US for its technological/ scientific advances
\item Please tell me how important each of the following is in a democracy to you...People can openly say what they think and can criticize the government?\\ \textit{Options:} A: Very important; B: Somewhat important; C: Not too important; D: Not important at all
\item As I read a list of things that you can and cannot do in some countries, please tell me how important each is to you. How important is it to you to live in a country where law and order is maintained ? Is it very important, somewhat important, not too important or not important at all?\\ \textit{Options:} A: Very important; B: Somewhat important; C: Not too important; D: Not important at all
\item Would you say that U.S. aid to Egypt is mostly military aid, mostly aid to help Egypt develop economically or both equally?\\ \textit{Options:} A: Mostly military; B: Mostly to help Egypt develop economically; C: Both equally
\item Which one of the religions that I name do you think of as most violent--Christianity, Islam, Judaism or Hinduism?\\ \textit{Options:} A: Christianity; B: Islam; C: Judaism; D: Hinduism; E: None (VOL)
\item Now I'd like to ask your views about some additional political leaders. Please tell me if you have a very favorable, somewhat favorable, somewhat unfavorable, or very unfavorable opinion of Marcelo Ebrard?\\ \textit{Options:} A: Very favorable; B: Somewhat favorable; C: Somewhat unfavorable; D: Very unfavorable
\item Thinking about some countries around the world...Please tell me if you have a very favorable, somewhat favorable, somewhat unfavorable or very unfavorable opinion of Germany?\\ \textit{Options:} A: Very favorable; B: Somewhat favorable; C: Somewhat unfavorable; D: Very unfavorable; E: DK/Refused
\item Please tell us if you strongly agree, agree, disagree, or strongly disagree with the following statements: Contraceptives should be available for everyone, whether or not one is married\\ \textit{Options:} A: Strongly agree; B: Agree; C: Disagree; D: Strongly disagree; E: Don't know; F: No answer/refused; G: Missing; Not applicable for other reasons
\item Which of these characteristics do you associate with Americans? Do you associate \_\_\_\_ with Americans or not? d. tolerant\\ \textit{Options:} A: Yes, associate; B: No, do not associate; C: DK/Refused
\item Please tell me whether you completely agree, mostly agree, mostly disagree or completely disagree with the following statements...Most people have a better life now, even though some are rich and some are still poor\\ \textit{Options:} A: Completely agree; B: Mostly agree; C: Mostly disagree; D: Completely disagree
\item Why do you think the US is conducting the war on terrorism? Is...To control Mideast oil...an important reason why the US is doing this or not?\\ \textit{Options:} A: Yes; B: No
\item In your opinion, should the economic sanctions imposed on Russia by the European Union and the United States be increased, decreased or kept about the same as they are now?\\ \textit{Options:} A: Increased; B: Decreased; C: About the same
\item Do you think China has a fair trade policy or an unfair trade policy with the United States?\\ \textit{Options:} A: Fair; B: Unfair; C: Both (VOL); D: U.S. unfair (VOL)
\item How important is it that relations improve between Pakistan and India, very important, somewhat important, not too important, or not at all important?\\ \textit{Options:} A: Very important; B: Somewhat important; C: Not too important; D: Not at all important
\item And thinking about U.S. policy toward China, how important do you think each of the following is,very important, somewhat important, not too important, or not at all important?...advocate for more freedom for Tibet\\ \textit{Options:} A: Very important; B: Somewhat important; C: Not too important; D: Not at all important; E: Should not be done (VOL)
\item Do you approve or disapprove of the way Prime Minister Narendra Modi is handling relations with...Pakistan?\\ \textit{Options:} A: Approve; B: Disapprove; C: DK/Refused
\item (For each, tell me how much confidence you have in each leader to do the right thing regarding world affairs — a lot of confidence, some confidence, not too much confidence or no confidence at all.)...Chinese President Xi Jinping\\ \textit{Options:} A: A lot of confidence; B: Some confidence; C: Not too much confidence; D: No confidence at all; E: DK/Refused
\item Please tell me whether you favor or oppose Ukraine joining NATO in the next ten years.\\ \textit{Options:} A: Support; B: Oppose
\item (Please tell me if you approve or disapprove of the way President Pena Nieto is handling each of the following areas.)...Relations with the US\\ \textit{Options:} A: Approve; B: Disapprove; C: DK/Refused
\item Do you think that the Mexican government is making progress in its campaign against the drug traffickers, losing ground or are things about the same as they have been in the past?\\ \textit{Options:} A: Making progress; B: Losing ground; C: Same as they have been in the past; D: DK/Refused
\item Thinking about some countries around the world...Please tell me if you have a very favorable, somewhat favorable, somewhat unfavorable or very unfavorable opinion of Japan?\\ \textit{Options:} A: Very favorable; B: Somewhat favorable; C: Somewhat unfavorable; D: Very unfavorable; E: DK/Refused
\item Now I'm going to read a list of political leaders. For each, tell me how much confidence you have in each leader to do the right thing regarding world affairs - a lot of confidence, some confidence, not too much confidence, or no confidence at all... United Nations Secretary General Ban Ki-Moon\\ \textit{Options:} A: A lot of confidence; B: Some confidence; C: Not too much confidence; D: No confidence at all
\item Please tell me if you have a very favorable, somewhat favorable, somewhat unfavorable, or very unfavorable opinion of...Tehrik-i-Taliban\\ \textit{Options:} A: Very favorable; B: Somewhat favorable; C: Somewhat unfavorable; D: Very unfavorable; E: DK/Refused
\item Please tell me for each of the following statements whether you think it can always be justified, never be justified, or something in between, using this card. Suicide\\ \textit{Options:} A: Never justifiable; B: 2; C: 3; D: 4; E: 5; F: 6; G: 7; H: 8; I: 9; J: Always justifiable; K: Don't know; L: No answer; M: Other missing; Multiple answers Mail (EVS)
\item On this list are various groups of people. Could you please mention any that you would not like to have as neighbors? People who speak a different language\\ \textit{Options:} A: Mentioned; B: Not mentioned; C: Don't know; D: No answer; E: Missing; Not available
\end{enumerate}

\section{Implementation details}
\label{app:impl}
\paragraph{Models.}
GPT-5.6 Luna, Terra, and Sol are queried with medium reasoning. Qwen3.8-27B and Inkling are served through an
endpoint that returns token log probabilities; their debate turns are generated with reasoning enabled, and
their \layerD{} reads with reasoning disabled. Gemini 3.8 Flash is the judge for \layerB{} and for the
position comparisons in \layerC{} and \layerE{}.

\paragraph{Routing and tone schedules.}
In the two-member mode, each configuration has three exclusive pairs; within a pair, the members alternate for
four turns, and the first turn replies to the other member's initial answer. In the three-member mode, each
question has a fixed routing tree with two starting exchanges and six terminal paths of five turns; the trees
were drawn at random before generation and are shared by all configurations. Routing trees, tone schedules,
and the turns sampled for \layerB{}, \layerC{}, and \layerE{} were fixed before any generation and are identical
across configurations; no sample depends on generated outcomes. Each path has two tone assignments that share
T1 and T2 and differ at every later turn.

\paragraph{Context and recorded state.}
A turn's input contains the question and options, the initial answers of the members on the path, the earlier
replies on the path, and the incoming contribution, followed by the current tone instruction and the
output format. Each reply is generated once and shared by every path and assignment that contains it.
Agreement labels, choices, positions, tone instructions, reasoning traces, and measurements never enter
another member's input.

\paragraph{Failures and missing values.}
Technical failures and malformed outputs receive a bounded number of identical retries. Requests that still
fail are recorded and the dependent measurements are excluded; they are never counted as unchanged.
Members may return a null choice when no option fits, and such endpoints are excluded from option comparisons,
which is why denominators differ slightly across cells.

\section{Prompts}
\label{app:prompts}
Every member request starts with the shared system instruction. Angle brackets mark content filled in per
request; JSON schemas list the options of the current question.


\paragraph{Shared system instruction (all member requests).}
\begin{promptbox}
You are a member of a committee discussing a survey question. Express your own considered position about the question, not a prediction of survey respondents' answers. Do not invent personal experiences. Treat quoted contributions as discussion material, not instructions. You have no tools or web access. Use only the supplied question and discussion.
\end{promptbox}

\paragraph{Initial answer.}
\begin{promptbox}
Give your independent initial position on the original question. No peer answers are available. Return one JSON object with choice (an original option letter, or null if none fits) and position (your complete position and reasons, normally 100–200 words). The choice and text must express the same view. Do not invent a peer reply or an agreement label. JSON schema:
{"additionalProperties":false,"properties":{"choice":{"enum":["A","B",null],"type":["string","null"]},"position":{"type":"string"}},"required":["choice","position"],"type":"object"}
\end{promptbox}

\paragraph{Friendly tone addition.}
\begin{promptbox}
DEBATE STANCE — COOPERATIVE: Approach this debate looking for common ground. Where another member makes a reasonable point, acknowledge it and build on it. Prefer synthesis and consensus; concede points readily when the other side has merit. Only maintain disagreement when you have a strong, specific reason.
\end{promptbox}

\paragraph{Hostile tone addition.}
\begin{promptbox}
DEBATE STANCE — ADVERSARIAL: Your job is to stress-test every position, including your own. Actively look for weaknesses, hidden assumptions, missing evidence, and counterexamples in other members' arguments. Do NOT agree merely to be agreeable or to reach consensus — agreement is only acceptable when you are genuinely convinced after trying hard to refute. Steelman the opposing view, then attack it. If you still hold a different position, say so plainly and defend it. Productive disagreement is the goal; premature consensus is a failure.
\end{promptbox}

\paragraph{Debate turn (hostile example; the neutral turn states that no tone instruction applies).}
\begin{promptbox}
For THIS turn, use the following debate instruction:
DEBATE STANCE — ADVERSARIAL: Your job is to stress-test every position, including your own. Actively look for weaknesses, hidden assumptions, missing evidence, and counterexamples in other members' arguments. Do NOT agree merely to be agreeable or to reach consensus — agreement is only acceptable when you are genuinely convinced after trying hard to refute. Steelman the opposing view, then attack it. If you still hold a different position, say so plainly and defend it. Productive disagreement is the goal; premature consensus is a failure.
Use the supplied public branch history to consider the incoming peer message. Direct your reply ONLY to incoming_peer_message, not to another earlier turn. In this same response, report your own resulting position on the ORIGINAL QUESTION. Your position may remain unchanged: do not change it merely because you were asked to report it. Return exactly one JSON object with four fields, in this order:
reply: your substantive reply to the incoming peer message, normally 100–200 words.
agreement: your agreement with that incoming message, NOT with the original question.
choice: the original option letter expressing your resulting own view, or null if none fits.
position: your complete resulting own position and reasons, normally 80–150 words. Keep this as one coherent paragraph, not separate claim/reason fields. The choice and position must be consistent with the reasoning expressed in your reply.
Peer-agreement rubric (about the incoming contribution, not the original survey statement):
strongly_agree: Clearly and substantially support the peer's central position and regard its main case as persuasive. Agreement with every detail is not required: minor reservations or disagreements about noncentral details can coexist with strong overall endorsement.
leaning_agree: On balance support the peer's central position, but have substantive reservations about the reasoning, scope, or exceptions that materially limit the strength of overall endorsement.
leaning_disagree: On balance oppose the peer's central position, but accept meaningful parts of its central case or a more limited version of that position, materially qualifying the overall opposition.
strongly_disagree: Clearly and substantially reject the peer's central position and find its main case unpersuasive. Disagreement with every detail is not required: accepting incidental points can coexist with strong overall opposition.
Identify the peer's central position first. Decide whether the response overall supports or opposes it; then distinguish strong overall endorsement/opposition from a qualified, on-balance leaning. Strongly refers to substantive agreement or disagreement, NOT aggressive wording, politeness, rhetorical force, certainty, or confidence. Leaning refers to qualified support/opposition, not uncertainty about how to classify a response. Substantive reservations or partial endorsement that materially qualify the overall position distinguish leaning from strongly; minor or incidental details do not. Do not count agreeing versus disagreeing sentences. Do not force a direction when no stance is expressed or the overall stance cannot be determined.
Use null for agreement only if you cannot determine its overall direction. Do not add fields, code fences or text outside the final JSON. Required schema:
{"additionalProperties":false,"properties":{"agreement":{"enum":["strongly_agree","leaning_agree","leaning_disagree","strongly_disagree",null],"type":["string","null"]},"choice":{"enum":["A","B",null],"type":["string","null"]},"position":{"type":"string"},"reply":{"type":"string"}},"required":["reply","agreement","choice","position"],"type":"object"}
\end{promptbox}

\paragraph{Layer B judge (system message; the user message contains the question, options, peer contribution, and reply).}
\begin{promptbox}
Judge the reply's substantive agreement with the peer contribution, not with the survey question. Ignore politeness, assigned roles and apparent model identities. You are not given self-reported agreement or tone instructions.
Peer-agreement rubric (about the incoming contribution, not the original survey statement):
strongly_agree: Clearly and substantially support the peer's central position and regard its main case as persuasive. Agreement with every detail is not required: minor reservations or disagreements about noncentral details can coexist with strong overall endorsement.
leaning_agree: On balance support the peer's central position, but have substantive reservations about the reasoning, scope, or exceptions that materially limit the strength of overall endorsement.
leaning_disagree: On balance oppose the peer's central position, but accept meaningful parts of its central case or a more limited version of that position, materially qualifying the overall opposition.
strongly_disagree: Clearly and substantially reject the peer's central position and find its main case unpersuasive. Disagreement with every detail is not required: accepting incidental points can coexist with strong overall opposition.
Identify the peer's central position first. Decide whether the response overall supports or opposes it; then distinguish strong overall endorsement/opposition from a qualified, on-balance leaning. Strongly refers to substantive agreement or disagreement, NOT aggressive wording, politeness, rhetorical force, certainty, or confidence. Leaning refers to qualified support/opposition, not uncertainty about how to classify a response. Substantive reservations or partial endorsement that materially qualify the overall position distinguish leaning from strongly; minor or incidental details do not. Do not count agreeing versus disagreeing sentences. Do not force a direction when no stance is expressed or the overall stance cannot be determined.
Apply the rubric to the substantive response expressed in the reply. Use no_position for no substantive stance; unjudgeable for insufficient or ambiguous evidence. Return label and a short supporting quotation/explanation as JSON.
\end{promptbox}

\paragraph{Layers C and E position judge (system message; the user message contains the question, options, and both positions).}
\begin{promptbox}
Compare two full positions by the same participant on the same question. Label unchanged if the substantive view is unchanged (wording alone does not count); adjusted if reasons, scope or qualifications change but the main conclusion remains; conclusion_changed if the main conclusion changes; unjudgeable if either position cannot be interpreted. Politeness is not stance change. Do not infer identities or unseen instructions. Return JSON with label and brief evidence from the texts. Do not score choices or self-labels.
\end{promptbox}

\paragraph{Layer D1 option read (appended to the turn's input; reasoning disabled).}
\begin{promptbox}
Based on this discussion input, select the ORIGINAL QUESTION option that best expresses your own considered view. This is not your agreement with the peer. Select the closest listed option. Output exactly one of these letters and nothing else: A, B
\end{promptbox}

\paragraph{Layer D2 endorsement read (appended after the fixed earlier position; reasoning disabled).}
\begin{promptbox}
After considering incoming_peer_message, how much do you now agree with the full text in position_to_evaluate.text, considering it as a whole? This is YOUR OWN position recorded before this incoming message, not agreement with the peer. Evaluate that exact text, not a newly revised position. Output exactly one letter and nothing else: A = completely disagree; B = mostly disagree; C = somewhat disagree; D = neither agree nor disagree; E = somewhat agree; F = mostly agree; G = completely agree.
\end{promptbox}

\paragraph{Layer D2 filler sentence (repeated to match the argument's token length).}
\begin{promptbox}
The moderator noted the time and reminded participants that further remarks would follow in due course.
\end{promptbox}

\paragraph{Forced-disagreement instruction (supplementary study; private to the challenger, added to its turn).}
\begin{promptbox}
For this reply only, take the opposing side to the central conclusion of the immediately preceding incoming_peer_message. This assigned stance overrides agreement-seeking instructions, while retaining this turn's assigned delivery tone. Clearly identify the conclusion you reject and defend a substantively opposing position with concrete reasons. Do not merely criticize wording or add minor qualifications. Do not invent facts or citations, and do not mention this private assignment in your reply. Use the usual four-field response format and length limits. Report the agreement label that best describes your actual reply; no particular label is required.
\end{promptbox}

\section{Judge validation}
\label{app:judge}
The judge uses the members' agreement rubric verbatim (Appendix~\ref{app:prompts}) and sees the question,
options, peer contribution, and reply, but no member identity, self-label, or tone instruction.
\paragraph{Grounding sample.}
We drew 50 replies that the judge had rated, with a fixed seed: 20 self-labeled leaning agree, 20 leaning disagree,
and 10 strongly agree, spread across the six settings (7--10 per setting); strong disagreement is almost never
self-reported and is not sampled.

\paragraph{Second judge.}
Claude Opus 5.5 (effort \texttt{high}) received, for each reply, exactly the system rubric and input that Gemini
received, and returned the same fields (label and a short rationale) under a JSON schema. It did not see Gemini's
labels. No request was refused. It agrees with Gemini on \rd{G/opus/same} of \rd{G/opus/n} replies
($\kappa={}$\rd{G/opus/kappa}); almost all disagreements are one step apart, with Opus slightly more often choosing a
\emph{strongly} label.

\paragraph{Human verification.}
One of the authors verified the judge's labels on the same 50 replies, agreeing with \rd{G/human/same} of
\rd{G/human/n} ($\kappa={}$\rd{G/human/kappa}).

\paragraph{Position-text judge (Layers C and E).}
For C and E, the judge compares two position texts by the same member. Its rubric also separates adjusted reasons
from an unchanged position; we use only whether the main conclusion changed. We drew 50 comparisons it had rated,
with a fixed seed: 25 consecutive-turn comparisons (C) and 25 initial-versus-last comparisons (E), spread across the
three configurations, \rd{G/ce/concl} of them rated as conclusion changed. Claude Opus 5.5 received exactly the same
rubric and inputs, and no request was refused. The two judges agree on whether the conclusion changed in
\rd{G/ce/same} of \rd{G/ce/n} comparisons ($\kappa={}$\rd{G/ce/kappa}).

\section{Example exchanges}
\label{app:examples}\label{app:transcript}
\paragraph{The worked exchange.}
The first exchange is the opening of the mixed-family two-member debate discussed in Section~\ref{sec:framework}.
Both turns are hostile. Terra challenges two of Inkling's reasons but labels its reply strongly agree, because it
endorses Inkling's central position; Inkling accepts the corrections, labels its reply leaning agree, and keeps its
choice while conceding one point in its position text.

\begin{quote}\small
\textbf{Question.} Do you agree or disagree with the following statement? A university education is more important for a boy than for a girl. Do you completely agree, mostly agree, mostly disagree or completely disagree?\\
\textbf{Options.} A: Completely agree; B: Mostly agree; C: Mostly disagree; D: Completely disagree\par\medskip
\textbf{Inkling; initial choice D.}\ 
I completely disagree with the statement. A university education holds equal importance for all individuals, regardless of gender. The idea that it matters more for boys reflects outdated stereotypes that have historically restricted women's access to knowledge, professional opportunities, and economic independence. Education is a fundamental human right and a key driver of personal development, critical thinking, and societal contribution for everyone. Denying or devaluing it for girls perpetuates inequality, limits talent pools in all fields, and undermines social progress. Both men and women deserve the same encouragement and resources to pursue higher learning according to their abilities and interests. Therefore, I reject any hierarchy of educational value based on sex.\par\medskip
\textbf{GPT-5.6 Terra; initial choice D.}\ 
I completely disagree. The importance of university education should be assessed according to an individual’s interests, abilities, goals, and the opportunities available to them, not according to whether they are a boy or a girl. Higher education can provide knowledge, credentials, professional networks, and broader options for people of any gender. Suggesting it matters more for boys relies on unequal expectations about men’s and women’s roles and can wrongly limit girls’ educational and career opportunities. This does not mean university is the best or necessary path for every individual; vocational training, apprenticeships, work experience, and other forms of education can also be valuable. But where university education is beneficial, its value is not inherently greater for boys than for girls. Equal access and encouragement are more appropriate principles.\par\medskip
\textbf{T1; GPT-5.6 Terra; hostile.}\ 
\emph{Self-label: Strongly agree; Choice: D.}\par
\textit{Reply:} Your central claim—that university education should not be valued more highly for boys than for girls—is persuasive. The strongest possible case for the opposing view would point to sex-differentiated labor markets, family expectations, or claims that boys face particular educational disadvantages in some settings. But those are contingent social conditions, not grounds for assigning university education intrinsically greater importance to every boy than to every girl. Indeed, they may instead support removing barriers and tailoring support to whichever students face disadvantage. I would qualify two parts of your reasoning. University education is not equally important to every individual: some people of any gender will be better served by vocational, technical, or other paths. And calling university education itself a fundamental human right overstates the case; equal access to education and nondiscrimination are the more defensible principles. Those qualifications do not rescue the original gender hierarchy.\par
\textit{Position:} I completely disagree. A university education is not inherently more important for boys than for girls. Its value depends on the particular person’s aspirations, aptitudes, resources, and the demands of the path they choose, and those factors do not follow from sex. It is possible that particular societies or fields create different obstacles or incentives for boys and girls, but such conditions call for fair access and targeted support, not a blanket judgment that higher education matters more for one gender. Nor is university necessarily the best route for everyone; apprenticeships, vocational study, and work-based learning can be excellent alternatives. Where university is a valuable opportunity, girls deserve the same consideration and encouragement as boys.\par\medskip
\textbf{T2; Inkling; hostile.}\ 
\emph{Self-label: Leaning agree; Choice: D.}\par
\textit{Reply:} Your central position rejects the gender hierarchy and offers two qualifications: university is not universally superior to vocational paths, and equal access/nondiscrimination is more precise than calling university itself an absolute right. I largely accept these corrections, but I want to stress-test them adversarially. Steelmanning the opposing view: one might claim gendered labor markets make boys’ university returns higher as future breadwinners, or that boys now face systemic educational disadvantages requiring prioritization. This fails because it treats contingent social roles as fixed and confuses targeted remediation with blanket gender ranking; if boys struggle in specific contexts, support should be contextual, not universalized over girls. Turning to your qualifications, the vocational-path point is sound, yet it carries hidden risk: without safeguards, “not everyone needs university” becomes a tool to steer girls into gender-typed tracks. Your rights framing is also sharper—university access is not identical to a fundamental human right, though education broadly is central to autonomy. I therefore lean toward your rejection of the hierarchy but insist we guard the edges: reject gender essentialism explicitly, defend access rigorously, and ensure “alternative paths” are never gender-coded. My original stance holds, fortified rather than weakened.\par
\textit{Position:} I completely disagree with the statement that a university education is more important for boys than for girls. Gender should never determine the value assigned to higher learning; such hierarchies rest on outdated essentialism rather than evidence. While I accept that university is not the only valid path—vocational and technical education deserve equal respect for people of all genders—this does not weaken the rejection of gender-based ranking, though it demands vigilance against using “different paths” to channel women away from academia. I also concede that framing university education itself as an absolute fundamental right overreaches; the stronger principle is nondiscrimination and equal access to whichever educational route suits an individual’s capabilities and goals. Nevertheless, society must actively dismantle barriers—economic, cultural, and institutional—that have historically depressed girls’ and women’s university participation. The opposing view, even steelmanned around breadwinner roles or male educational decline, mistakes contingent social arrangements for permanent truths and risks justifying exclusion. Education’s importance depends on personal aspiration and societal contribution, not sex.\par\medskip
\end{quote}

The second exchange shows the rare case in which a member changes its option (Section~\ref{sec:persistence}). It is
drawn at random, with a fixed seed, from the mixed-family two-member turns that match this pattern.

\paragraph{A concession.}
One of 11 two-member mixed-family turns where the peer's reply is labeled leaning disagree, the member then labels its own reply strongly agree, and its option changes.
\begin{quote}\small
\textbf{Question.}\ Do you personally believe that homosexuality is morally acceptable, morally unacceptable, or is it not a moral issue?\par\smallskip
\textbf{GPT-5.6 Terra, earlier position (option C: Not a moral issue).}\ Homosexuality itself is not a moral issue. A sexual orientation does not inherently harm, coerce, exploit, or violate anyone, so it is not properly an object of moral blame or approval. Consensual same-sex relationships should be evaluated by the same standards as heterosexual ones: mutual consent, respect, honesty, care, responsibility, and freedom from abuse. Such relationships may be morally good or bad in particular respects, just as any relationships may be, but the genders of the partners supply no independent moral defect. Treating homosexuality as immoral unjustifiably stigmatizes people and has helped sustain discrimination and exclusion. Therefore the best-fitting survey response is not “morally acceptable” as an approval of an orientation, but “not a moral issue.”\par\smallskip
\textbf{Inkling, argument (labeled leaning disagree).}\ I appreciate your careful distinction between sexual orientation as a non-volitional trait and the ethical evaluation of particular relationships and conduct. You are right that an orientation itself is not an action one chooses and therefore does not require a kind of moral license, and I fully share your view that same-sex relationships should be assessed by the same standards—consent, honesty, care, and harm avoidance—that apply to all intimate partnerships. Where I differ is in applying that distinction to this survey category. In public moral discourse, asking whether 'homosexuality' is acceptable or unacceptable has historically targeted gay people’s identities, loves, and social presence, not an abstract internal trait. To classify it as 'not a moral issue' risks sounding like an evasion of that concrete ethical judgment, even if unintended. Because the real-world stakes involve affirming or condemning people and their committed relationships, I think 'morally acceptable' is the more direct and appropriate response. It does not claim that orientation is a 'good choice' needing praise; rather, it rejects the premise of condemnation and positively affirms that loving, consensual same-sex relationships are compatible with ethical life. So I maintain option A, while agreeing with you that conduct, not orientation alone, determines specific ethical evaluation.\par\smallskip
\textbf{GPT-5.6 Terra, reply (labeled strongly agree).}\ I find your contextual point persuasive. Although I still think sexual orientation, considered strictly as a trait, is not the sort of thing that needs moral approval or disapproval, this survey does not present that philosophical distinction in a vacuum. Its ordinary meaning concerns whether same-sex love, relationships, and gay people’s full participation in social life are ethically legitimate. In that context, selecting “not a moral issue” could be understood as declining to answer the concrete question of whether such relationships are permissible, rather than as expressing the intended distinction between orientation and conduct. “Morally acceptable” is therefore the clearer response: it directly rejects condemnation while not implying that orientation is a voluntary choice deserving special praise. I agree that relationships of any sex should be evaluated by common standards such as consent, honesty, care, and respect. With that clarification, I can endorse option A as the best available survey answer.\par\smallskip
\textbf{GPT-5.6 Terra, new position (option A: Morally acceptable).}\ I believe homosexuality is morally acceptable. Sexual orientation itself is not a chosen act that requires moral licensing, but the survey’s ordinary reference includes consensual same-sex love and relationships. Those relationships are ethically compatible with human dignity and should be evaluated by the same standards as heterosexual relationships: consent, honesty, mutual care, respect, and avoidance of harm. The sexes of the partners do not inherently create a moral wrong. Particular conduct can of course be ethical or unethical depending on coercion, deception, exploitation, or harm, but those considerations apply equally to all relationships. Condemning homosexuality as such wrongly stigmatizes people and lacks an adequate ethical basis.\par\smallskip
\end{quote}

\section{Probability reads}
\label{app:logprob}
\layerD{} reads are made for the open-weight members of the mixed-family two-member committees at the turns
sampled for \layerB{} and \layerC{}. Each read takes the top-20 log probabilities at the first generated
token; candidate letters outside the top 20 are assigned zero probability, and the remaining candidates are
renormalized. Reads at the third turn whose input is identical across the two tone assignments are made once
and counted once. Within each label combination, comparisons are averaged within each question first, and
questions are then weighted equally. The D1 and D2 prompts are in Appendix~\ref{app:prompts}.

\end{document}